\PassOptionsToPackage{usenames,dvipsnames}{xcolor}
\documentclass[preprint,12pt]{elsarticle}
\usepackage{comment}
\usepackage{url} 
\usepackage{soul}
\usepackage[utf8]{inputenc}
\usepackage[export]{adjustbox}
\usepackage{amsmath, amsfonts, amssymb, amsthm, mathtools}
\usepackage{algpseudocode}
\usepackage{natbib}
\usepackage{booktabs, tabu, multicol, multirow, makecell, bigstrut}
\usepackage{caption}
\usepackage[subrefformat=parens]{subcaption}
\usepackage{hyperref}
\usepackage[capitalise]{cleveref}
\usepackage{xstring}
\usepackage{tikz, pgfplots, pgfplotstable}

\usetikzlibrary{calc, matrix}
\usepgfplotslibrary{groupplots, statistics, fillbetween}
\definecolor{seagreen}{RGB}{46, 139, 87}
\definecolor{midnight}{RGB}{25, 25, 112}
\definecolor{orange}{RGB}{255, 165, 0}

\tikzset{
    every picture/.style = {
        very thick,
        draw = midnight,
        fill = midnight,
        node distance = 0pt,
        auto
    },
    every node/.style = {
        inner sep = 0.5ex,
        outer sep = 0pt,
        minimum width = 3ex,
        minimum height = 4ex
    },
    vertex/.style = {
        draw,
        circle,
        minimum size = 5pt
    },
    arrow/.style = {
        ->, >=stealth,
        auto,
        draw = midnight
    },
    var/.style = {
        font = \vphantom{$\hat{A}g$}
    },
}

\pgfplotsset{
    compat=newest,
    every axis/.append style={
		thick,
		line width=1pt,
		x tick label style={
		    /pgf/number format/.cd,
		    set thousands separator={}
		}
	},
	all plot/.style={
        width=0.85*\textwidth,
        height=0.3*\textheight,
        xtick=data,
	},
	error plot/.style={
        all plot,
        cycle list={
            Maroon,every mark/.append style={fill=Maroon,solid},mark=*\\%
            MidnightBlue,every mark/.append style={fill=MidnightBlue,solid},mark=square*\\%
        },
        xlabel={Number of KL terms},
        ylabel={$\ell_2$ errors},
	},
	time plot/.style={
	    all plot,
	    cycle list={
	        MidnightBlue,every mark/.append style={fill=MidnightBlue,solid},mark=square*\\%
	    },
	    ylabel={execution time (s)},
	},
    trend plot/.style={
        node distance=-.5pt,
        width=0.65*\textwidth,
        height=0.35*\textheight,
        xtick=data,
        cycle multi list={
            {Maroon, mark=square*, draw=none}, {ForestGreen, mark=*}, {MidnightBlue, mark=diamond*}, {BurntOrange, mark=triangle*}, {CarnationPink, mark=*}\nextlist
            only marks, mark=none
        },
        xlabel={Number of FV cells},
        ylabel={Execution time (s)},
        xticklabel={
            \pgfkeys{/pgf/fpu=true}
            \pgfmathparse{exp(\tick)}
            \pgfmathprintnumber[fixed relative, precision=4]{\pgfmathresult}
            \pgfkeys{/pgf/fpu=false}
        },
        legend cell align=left,
        legend pos=outer north east,
        legend style={
            font=\small,
            draw=none,
        },
    },
    line and fill/.style={
		legend image code/.code={%
			\fill [##1,opacity=0.1,draw=none]
				(0mm,-1ex) rectangle (6mm,1ex);
			\draw [##1] (0mm,0mm) -- plot (3mm,0mm) -- (6mm,0mm);
		}
	},
}

\newcommand{\colorbar}[3]{
\begin{tikzpicture}[baseline={(0, -2.75em)}]
    \pgfplotscolorbardrawstandalone[
        parent axis height/.initial = 5em,
        point meta min=#1,
        point meta max=#2,
        colormap/jet,
        colorbar,
        colorbar style={
            width=0.4em,
            ytick=#3
    }]
\end{tikzpicture}
}

\newcommand{\errorband}[6][]{
	\addplot[name path={#2_#4},draw=none,forget plot]
		table [x={#3},y={#4}] {#2};
	\addplot[name path={#2_#5},draw=none,forget plot]
		table [x={#3},y={#5}] {#2};
	\addplot[#1,opacity=0.1,draw=none,forget plot]
		fill between[of={#2_#4} and {#2_#5}];
	\addplot[#1,line width=1pt,fill=none,forget plot]
		table [x={#3},y={#6}] {#2};
	\addlegendimage{line and fill,#1};
}

\newcommand\setv[3]{\expandafter\xdef\csname #1_#2\endcsname{#3}}
\newcommand\getv[2]{\csname #1_#2\endcsname}

\begin{document}

\begin{frontmatter}
  \journal{Journal of Computational Physics}

\title{Gaussian process regression and conditional Karhunen-Lo\'{e}ve models for data assimilation in inverse problems\tnoteref{t1}}%
\tnotetext[t1]{%
  This research was partially supported by the U.S. Department of Energy (DOE) Advanced Scientific Computing program. Pacific Northwest National Laboratory is operated by Battelle for the DOE under Contract DE-AC05-76RL01830.%
}

\author[1]{Yu-Hong Yeung}%
\ead{Yu-Hong.Yeung@pnnl.gov}

\author[1]{David A. Barajas-Solano}%
\ead{David.Barajas-Solano@pnnl.gov}
  
\author[1,2]{Alexandre M. Tartakovsky\corref{cora}}%
\ead{amt1998@illinois.edu}

\cortext[cora]{Corresponding author}

\affiliation[1]{%
  organization={Physical and Computational Sciences Directorate, Pacific Northwest National Laboratory},%
  city={Richland},%
  postcode={99354},%
  state={WA},%
  country={USA}%
}

\affiliation[2]{%
  organization={Department of Civil and Environmental Engineering, University of Illinois Urbana-Champaign},%
  city={Urbana},%
  postcode={61801},%
  state={IL},%
  country={USA}%
}

\begin{abstract}
  We present a model inversion algorithm, CKLEMAP, for data assimilation and parameter estimation in partial differential equation models of physical systems with spatially heterogeneous parameter fields.
  These fields are approximated using low-dimensional \emph{conditional Karhunen-Lo\'{e}ve expansions} (CKLEs), which are constructed using Gaussian process regression (GPR) models of these fields trained on the parameters' measurements. 
  We then assimilate measurements of the state of the system and compute the \emph{maximum a posteriori} (MAP) estimate of the CKLE coefficients by solving a nonlinear least-squares problem.
  When solving this optimization problem, we efficiently compute the Jacobian of the vector objective by exploiting the sparsity structure of the linear system of equations associated with the forward solution of the physics problem.
  
  The CKLEMAP method provides better scalability compared to the standard MAP method.
  In the MAP method, the number of unknowns to be estimated is equal to the number of elements in the numerical forward model.
  On the other hand, in CKLEMAP, the number of unknowns (CKLE coefficients) is controlled by the smoothness of the parameter field and the number of measurements, and is in general much smaller than the number of discretization nodes, which leads to a significant reduction of computational cost with respect to the standard MAP method.
  To show this advantage in scalability, we apply CKLEMAP to estimate the transmissivity field in a two-dimensional steady-state subsurface flow model of the Hanford Site by assimilating synthetic measurements of transmissivity and hydraulic head.
  We find that the execution time of CKLEMAP scales nearly linearly as $N^{1.33}$, where $N$ is the number of discretization nodes, while the execution time of standard MAP scales as $N^{2.91}$.
  The CKLEMAP method improved execution time without sacrificing accuracy when compared to the standard MAP method.
\end{abstract}

\begin{highlights}
\item We propose CKLEMAP as an efficient alternative to the maximum a posteriori probability (MAP) method of parameter estimation for partial differential equations.
\item The efficiency is due to the use of a conditional Karhunen-Lo\'{e}ve representation of the parameter field and an acceleration scheme for Jacobian computations. 
\item CKLEMAP and MAP scale as $N^{1.3}$ and $N^3$, where $N$ is the number of nodes of degrees of freedom in the discretization of the governing partial differential equation.
\item CKLEMAP is as accurate as MAP but significantly faster for large-scale parameter estimation problems.
\end{highlights}



\end{frontmatter}


\section{Introduction}\label{sec:intro}

Parameter estimation is a critical part of developing partial differential equation (PDE) models of natural or engineered systems. 
In heterogeneous systems, parameters vary in space (and, possibly, time), and the destructive nature and high cost of collecting measurements limit the number of direct parameter measurements that can be gathered.
As a consequence, modelers are tasked with solving the inverse problem, i.e.,  estimating parameters from a limited number of direct measurements and, usually, a larger number of indirect measurements, e.g., measurements of the states in the PDE model. 
In the context of subsurface flow and transport, such observables include hydraulic head and tracer breakthrough measurements at observation wells, among others.

The heterogeneity of parameters gives rise to two challenges: (1) spatial heterogeneity must be \emph{parameterized}, either naively, using the grid discretization of the PDE's domain, or through some other scheme; and (2) sparse-in-space measurements are often not enough to fully characterize spatial heterogeneity, thus it is necessary to introduce assumptions about spatial heterogeneity that regularize the inverse problem.

Once parameterization and regularization schemes have been selected, one can compute the \emph{maximum a posteriori} (MAP) estimate of the model parameters.
The MAP estimate is computed by solving a PDE-constrained optimization problem consisting of minimizing a certain norm of the difference between predicted and measured observables (data misfit term) plus a regularizing penalty.
Assuming that the solution is obtained at a global minimum, the MAP estimate is equivalent to the largest mode of the Bayesian posterior with the data misfit term corresponding to the (negative) Bayesian log-likelihood and the regularizing penalty corresponding to the (negative) Bayesian log-prior~\citep{kitanidis-1996-geostatistical,stuart-2010-inverse,barajassolano-2014-linear}.
One can drop the PDE constraint by modeling the predicted observables via a ``surrogate'' model, at the cost of constructing said model either on the fly (e.g.,~\cite{wild-pounders}) or ahead of tackling the inverse problem (e.g.,~\cite{mo-2020-integration,mo-2019-deepconv,mo-2019-deepauto}).
Alternatives to MAP estimation for nonlinear problems include iterative linear filtering and smoothing~\citep{chada-2021-iterative,zhang-2018-ilues}.
In this work, by ``MAP method'' we will refer to MAP estimation via nonlinear least-squares using the parameterization in terms of the degrees of freedom of the spatial grid discretization of the forward solver scheme.

The pilot point method (PPM)~\citep{certes-1991-application,doherty-2010-approaches,tonkin-2005-hybrid} provides parameterization and regularization by modeling parameter fields as a regressor computed from a set of spatially discrete values (``pilot points'') of the parameter fields.
These pilot points then become the parameters to be estimated via PDE-constrained optimization.
The choice of the number and locations of pilot points is not trivial and significantly affects the quality and time-to-solution of the inverse problems.
To address these challenges,~\cite{tonkin-2005-hybrid} proposed to use the singular value decomposition of the sensitivities of observables with respect to the pilot points to reduce the effective dimension of the pilot point parameterization.
Beyond PPM, other parameterizations and regularization schemes have been proposed.
For example, \cite{xu-2021-CMAME} represented the parameter field with a deep neural network and~\cite{omalley-2019-learning,mo-2020-integration} used the latent space representation of the parameter fields defined by a variational autoencoder and a convolutional adversarial autoencoder, respectively.

Scientific machine learning (SciML) algorithms provide both an alternative and a complement to the PDE-constrained optimization and linear filtering-based approaches to inverse problems described above.
SciML approaches for inverse problems can be roughly classified into two families: physics-informed deep learning (DL) and DL for constructing surrogate models.
In physics-informed DL methods~\citep{tartakovsky-2020-pinn,he-2020-pinn,yang-2019-highly}, the parameters and states of PDE models are represented by DL models such as feed-forward or convolutional neural networks; then, the parameters of these DL models are estimated by minimizing an objective consisting of the data misfit term plus a weighted penalty on the PDE model residuals evaluated at certain points in the simulation domain.
This objective corresponds to the so-called ``penalty'' approximation of the corresponding constrained minimization problem with a fixed penalty weight~\cite{leeuwen-2015-penalty}.
The physics-informed DL approaches rely on the expressive capacity of DL models to accurately represent parameters and states.
On the other hand, DL surrogate modeling approaches use DL models to approximate the map from parameters to observables~\citep{mo-2020-integration,mo-2019-deepconv,mo-2019-deepauto,kadeethum-2021-framework}.
These approaches rely on the capacity of DL models to approximate functions of high-dimensional inputs.
Other recent developments include ``neural operator'' methods,
which aim to learn the PDE solution as an explicit function of the model parameters~\cite{lu-2021-nature}.

Karhunen-Lo\`{e}ve expansions (KLEs) are extensively employed to parameterize spatially heterogeneous fields for both uncertainty quantification and model inversion tasks.
In~\citep{tipireddy-2020-conditional}, the \emph{conditional} KLE of the parameter field was conditioned on the direct field's measurements, leading to \emph{conditional} KL expansions (CKLEs).
It was demonstrated that using CKLEs instead of KLEs reduces the variance of the stochastic model of the parameter field and reduces uncertainty in the forward models. 
In~\citep{barajassolano-2019-pickle,yeung-2022-physics},  CKLEs were used to represent both parameter and state fields for solving inverse problems. The CKLE parameters were estimated by minimizing the residuals of the governing equations.  
The resulting ``physics-informed CKLE'' algorithm (PICKLE) was shown to provide approximate solutions to the inverse problem of accuracy comparable to PDE-constrained optimization-based methods but at a significantly lower computational cost.

Here, we propose solving inverse problems in PDE models by representing the parameter fields using CKLEs conditioned on available direct measurements of these fields and then estimating the CKLE coefficients via nonlinear least-squares.
We refer to this combination of MAP estimation and CKLEs as ``CKLEMAP.''
Compared to PICKLE, CKLEMAP is free of the errors introduced by the approximation of the state with the CKLE expansions and the penalty approximation of the PDE constraint, which leads to more accurate solutions to the inverse problem at the cost of having to solve the forward problem during the nonlinear least-squares minimization procedure.
Nevertheless, we significantly reduce the execution time of model inversion with respect to the MAP method by drastically reducing the number of parameters to be estimated.
We note that while KLEs, and more generally the spectrum of Gaussian process covariance models, have been extensively used to parameterize heterogeneous fields in Bayesian parameter estimation (e.g.,~\cite{kitanidis-2014-principal,lee-2014-largescale,markouk-2009-dimensionality,mo-2019-deepconv}), the application of KLE in deterministic inverse methods has not been explored and is the subject of our work.
Furthermore, we demonstrate the advantage of using the CKLE representation as opposed to the one based on KLE\@.

We apply CKLEMAP to a high-dimensional (approximately 1000 parameters in the CKLE are needed to accurately represent  the transmissivity field) stationary groundwater flow model of the Hanford Site, a former nuclear production complex on the west shore of the Columbia River in the Columbia Basin in the southeast part of the state of Washington in the United States and currently operated by the United States Department of Energy. 
We use CKLEMAP to estimate the transmissivity field from synthetic measurements of the transmissivity and hydraulic head fields.
These measurements are generated using the hydraulic conductivity measurements and boundary conditions obtained in the Hanford Site calibration study~\cite{cole-2001-transient}.

We compare the CKLEMAP and MAP methods and find that both methods are very close in accuracy with respect to the reference field.
On the other hand, we find that the computational cost of MAP increases with the problem size (the number $N$ of finite volume cells) as $N^{2.91}$, while the cost of CKLEMAP increases as $N^{1.33}$.
We also observe that for $N = 5900$, the execution time of CKLEMAP is one order of magnitude smaller than that of MAP, and for $N_{FV}=23600$, we estimate that CKLEMAP would be more than two orders of magnitude faster than MAP (the execution time of CKLEMAP is found to be $\approx 8\times 10^2$ s, and the execution time of MAP of approximately $2 \times 10^5$ s is estimated from the scaling relationship).
The choice of synthetic (as opposed to the field) measurements of the hydraulic head allows us to have a reference transmissivity field for comparing the accuracy of the MAP and CKLEMAP methods while preserving the complexity of boundary conditions and the transmissivity field of the Hanford Site.

\section{Groundwater flow model}\label{sec:problem}

We consider two-dimensional flow in a heterogeneous porous medium in the domain $D \subset \mathbb{R}^2$. 
Given some sparse measurements of the transmissivity $T(x)\colon D \to \mathbb{R}^+$ and the hydraulic head $u(x) \colon D \to \mathbb{R}$, our goal is to estimate the spatial distribution of transmissivity.
Flow in porous media is described by the boundary value problem (BVP)
\begin{align}
  \label{eq:pde}
  \nabla \cdot \left [ T(x) \nabla u(x) \right ] & = 0, && x \in D,\\
  \label{eq:pde-flux-bc}
  T(x) \nabla u(x) \cdot \vec{n}(x) &= -q_\mathcal{N}(x), && x \in \Gamma_\mathcal{N},\\
  \label{eq:pde-head-bc}
  u(x) &= u_\mathcal{D}(x), && x \in \Gamma_\mathcal{D},
\end{align}
where $\Gamma_\mathcal{N}$ and $\Gamma_\mathcal{D}$ are the disjoint subsets of the boundary of the domain $D$, where the Neumann and Dirichlet boundary conditions (BCs) are prescribed, respectively.
The flux $q_\mathcal{N} \in \mathbb{R}$ at the Neumann boundary $\Gamma_\mathcal{N}$ is  in the direction of the outward-pointing unit vector $\vec{n} \in \mathbb{R}^2$ normal to $\Gamma_\mathcal{N}$.
The prescribed hydraulic head at $\Gamma_{\mathcal{D}}$ is denoted as $u_\mathcal{D} \in \mathbb{R}$.

In groundwater models, Dirichlet BCs describe water levels in the lakes and rivers connected to the aquifer.
Since it is possible to measure the water levels relatively accurately, we treat the Dirichlet boundary conditions as deterministic.
Furthermore, we assume that the homogeneous Neumann boundary condition ($q_\mathcal{N}=0$) is imposed over the subset of $\Gamma_{\mathcal{N}}$ formed by the impermeable boundaries of the aquifer.
The rest of $\Gamma_{\mathcal{N}}$ is assumed to be formed by recharge areas where the values of $q_\mathcal{N} > 0$. The boundary fluxes from recharge areas are difficult to measure; therefore, we treat the non-zero fluxes as random variables and estimate them along with the transmissivity field $T$ as part of the inverse solution. 

The MAP method (described in detail in Section~\ref{sec:MAP}) requires solving the governing equation for different BCs and realizations of $T$, which in general must be done numerically.    
In this study, we solve the governing equation using a cell-centered finite volume (FV) scheme with $N$ quadrilateral cells, and the fluxes across cell faces are approximated using the two-point flux approximation (TPFA).
For simplicity, we assume that $\Gamma_{\mathcal{N}}$ and $\Gamma_{\mathcal{D}}$ are entirely composed of cell faces.
Let $\hat{x}_i$ denote the $i$th cell center, with $i \in [1, N]$.
We denote by $u_i \equiv u(\hat{x}_i)$ and $y_i \equiv y(\hat{x}_i)$ the discrete values of the hydraulic head field $u$ and log-transmissivity $y \equiv \ln T$ field evaluated at the $i$th FV cell centers.
These discrete values are organized into the column vectors $\mathbf{u} \equiv [u_1, \dots, u_N]^{\top} \in \mathbb{R}^N$ and $\mathbf{y} \equiv [y_1, \dots, y_N]^{\top} \in \mathbb{R}^N$, respectively.

Then, the FV-TPFA discretization of the BVP~\labelcref{eq:pde,eq:pde-flux-bc,eq:pde-head-bc} yields the system of equations linear in $\mathbf{u}$,
\begin{equation}
  \label{eq:pde-discretized}
  \mathbf{l}(\mathbf{u}, \mathbf{y}) \equiv \mathbf{A}(\mathbf{y}) \mathbf{u} - \mathbf{b}(\mathbf{y}) = 0,
\end{equation}
with stiffness matrix $\mathbf{A} \colon \mathbb{R}^N \to \mathbb{R}^{N \times N}$ and right-hand vector side $\mathbf{b} \colon \mathbb{R}^N \to \mathbb{R}^N$.
Here, $\mathbf{l} \colon \mathbb{R}^N \times \mathbb{R}^N \to \mathbb{R}^N$ denotes the vector of discretized BVP residuals whose entries correspond to the mass balance for each FV cell.
The set of FV cells $\mathcal{C}$ can be partitioned into three subsets: $\mathcal{N}$, the set $N_{\mathcal{N}}$ of cells adjacent to $\Gamma_\mathcal{N}$, $\mathcal{D}$, the set $N_{\mathcal{D}}$ of cells adjacent to $\Gamma_\mathcal{D}$, and the set of ``interior'' cells $\mathcal{I} = \mathcal{C} \setminus (\mathcal{D} \cup \mathcal{N})$ (that is, the cells to which boundary conditions do not contribute directly to their mass balance).
The set $\mathcal{I}$ has cardinality $N_{\mathcal{I}} = N - N_{\mathcal{N}} - N_{\mathcal{D}}$.

\section{MAP formulation}\label{sec:MAP}

We assume that $N_{\mathbf{u}_{\mathrm{s}}}$ and $N_{\mathbf{y}_{\mathrm{s}}}$ measurements of $\mathbf{u}$ and $\mathbf{y}$, denoted by $\mathbf{u}_{\mathrm{s}}$ and $\mathbf{y}_{\mathrm{s}}$, respectively, are collected at the cell centers indicated by the vectors of observation indices $\mathcal{I}_u$ and $\mathcal{I}_y$, respectively.
That is,
\begin{equation*}
  [ \mathbf{u}_{\mathrm{s}} ]_i \equiv u(\hat{x}_{[ \mathcal{I}_u ]_i}), \quad [ \mathbf{y}_{\mathrm{s}} ]_i \equiv y(\hat{x}_{[ \mathcal{I}_u ]_j}), \quad i \in [1, N_{\mathbf{u}_{\mathrm{s}}}], \ j \in [1, N_{\mathbf{y}_{\mathrm{s}}}].
\end{equation*}
Using these measurements, we aim to estimate $\mathbf{y}$.

The MAP estimator~\citep{kitanidis-1996-geostatistical} of $\mathbf{y}$ is computed by minimizing the sum of the $\ell_2$-norm of the discrepancy between measurements and model predictions, plus a regularization penalty on $\mathbf{y}$, that is, by solving the PDE-constrained minimization problem
\begin{equation}
  \label{eq:pde-constrained-opt-reg-general}
  \begin{aligned}
    \min_{\mathbf{u}, \mathbf{y}} \quad & \frac{1}{2} \| \mathbf{u}_{\mathrm{s}} - \mathbf{H}_{\mathbf{u}} \mathbf{u} \|^2_2 + \frac{1}{2} \| \mathbf{y}_{\mathrm{s}} - \mathbf{H}_{\mathbf{y}} \mathbf{y} \|^2_2 + \gamma \mathcal{R}(\mathbf{y}), \\
    \text{s.t.} \quad & \mathbf{l}(\mathbf{u}, \mathbf{y}) = 0,
  \end{aligned}
\end{equation}
where $\mathcal{R}(\mathbf{y})$ is the regularization penalty, $\gamma > 0$ is a regularization weight, and $\mathbf{H}_{\mathbf{u}} \colon \mathbb{R}^{N_{u_s} \times N}$ and $\mathbf{H}_{\mathbf{y}} \colon \mathbb{R}^{N_{y_s} \times N}$ are observation matrices, which downsample $\mathbf{u}$ and $\mathbf{y}$ using the observation indices $\mathcal{I}_u$ and $\mathcal{I}_y$, respectively.
Specifically, $\mathbf{H}_{\mathbf{u}} \equiv \mathbf{I}_N [\mathcal{I}_u]$, and $\mathbf{H}_{\mathbf{y}} \equiv \mathbf{I}_N [\mathcal{I}_y]$ are submatrices of the $N \times N$ identity matrix $\mathbf{I}_N$ corresponding to the rows of indices $\mathcal{I}_u$ and $\mathcal{I}_y$, respectively.

For $\mathbf{y}$, we employ the so-called ``$H^1$ regularization,'' which penalizes the $H^1$ seminorm of $\mathbf{y}$ (the $\ell_2$-norm of the gradient of $\mathbf{y}$).
In the discrete case, the $H^1$ seminorm penalty is of the form $\| \mathbf{D} \mathbf{y} \|^2_2$, where $\mathbf{D}$ is the TPFA discretization of the gradient operator such that $\mathbf{D} \mathbf{y}$ is equal to the gradients of $\mathbf{y}$ across the interior faces of the FV discretization.
The resulting PDE-constrained minimization reads
\begin{equation}
  \label{eq:pde-constrained-opt-reg}
  \begin{aligned}
    \min_{\mathbf{u}, \mathbf{y}} \quad & \frac{1}{2} \| \mathbf{u}_{\mathrm{s}} - \mathbf{H}_{\mathbf{u}} \mathbf{u} \|^2_2 + \frac{1}{2} \| \mathbf{y}_{\mathrm{s}} - \mathbf{H}_{\mathbf{y}} \mathbf{y} \|^2_2 + \frac{\gamma}{2} \| \mathbf{D} \mathbf{y} \|^2_2, \\
    \text{s.t.} \quad & \mathbf{l}(\mathbf{u}, \mathbf{y}) = 0,
  \end{aligned}
\end{equation}

The MAP estimates $\hat{\mathbf{u}}$ and $\hat{\mathbf{y}}$ obtained from~\cref{eq:pde-constrained-opt-reg} are equivalent to the largest mode $(\hat{\mathbf{u}}, \hat{\mathbf{y}})$ of the joint posterior distribution of $(\mathbf{u}, \mathbf{y})$ in a Bayesian interpretation of the inverse problem, in which the data misfit terms correspond to a Gaussian negative log-likelihood and the regularization penalty to a Gaussian negative log-prior.

\section{CKLEMAP method for inverse problems}\label{sec:cklemap}

\subsection{Parameterizing \texorpdfstring{$y(x)$}{y(x)} via conditional Karhunen-Lo\'{e}ve expansions}

As in the PICKLE method~\cite{ barajassolano-2019-pickle,yeung-2022-physics}, we represent the unknown parameter field $y(x)$ using the truncated CKLE 
\begin{equation}
  \label{eq:ckle_y}
  y^c(x, \boldsymbol{\xi}) \equiv \bar{y}^c(x) + \sum_{i=1}^{N_y} \phi_i^y(x) \sqrt{\lambda_i^y} \xi_i,
\end{equation}
where $\boldsymbol{\xi} \equiv (\xi_1, \xi_2, \ldots, \xi_{N_y})^\top$ is the vector of CKLE coefficients and the eigenpairs $\{\phi_i^y(x), \lambda_i^y\}_{i=1}^{N_y}$ are the solutions of the eigenvalue problem
\begin{equation}
  \label{eigenproblems}
  \int_D C^c_y(x, x') \phi^y(x') \, \mathrm{d} x' = \lambda^y\phi^y(x).
\end{equation}
Here, $\bar{y}^c(x)$ and $C^c_y(x, x')$ denote the mean and covariance of $y(x)$ conditioned on the measurements $\mathbf{y}_c$.

The CKLE is truncated (i.e., $N_y$ is selected) such as to achieve a desired relative tolerance  
\begin{equation}
\text{rtol}_y \equiv \sum^{N}_{i = N_y + 1} \lambda_i^y / \sum^{N}_{i = 1} \lambda_i^y,
\end{equation}
where $N$ is the number of FV cells.



The GPR (or Kriging) equations are used to compute $\overline{y}^c(x)$ and  $C^c_y(x, y)$: 
\begin{align}
  \label{eq:gpr-mean}
  \bar{y}^c(x) &=  \mathbf{C}(x) \mathbf{C}^{-1}_{\mathrm{s}}  \mathbf{y}_{\mathrm{s}},\\
  \label{eq:gpr-var}
  C^c_y(x, x') &= C_y(x, x') - \mathbf{C}(x) \mathbf{C}^{-1}_{\mathrm{s}} \mathbf{C}(x'),
\end{align}
where $\mathbf{C}_{\mathrm{s}}$ is the $N_{\mathbf{y}_{\mathrm{s}}} \times N_{\mathbf{y}_{\mathrm{s}}}$ observation covariance matrix with elements $[ \mathbf{C}_{\mathrm{s}} ]_{ij} = C_y(\hat{x}_{[\mathcal{I}_y]_i}, \hat{x}_{[\mathcal{I}_y]_j})$ and $\mathbf{C}(x)$ is the  $N_{\mathbf{y}_{\mathrm{s}}}$-dimensional vector function with components $[ \mathbf{C} (x) ]_i = C_y(x, \hat{x}_{[ \mathcal{I}_y ]_i})$.

The prior covariance kernel $C_y(x, y)$ is estimated as in the GPR method by choosing a parameterized covariance model and computing its hyperparameters by minimizing the marginal log-likelihood of the data $\mathbf{y}_{\mathrm{s}}$~\citep{rasmussen-2003-gaussian}.
In this work, we employ the $5/2$-Mat\'{e}rn kernel as the prior covariance model,
\begin{equation*}
  C_y(x,y) = \sigma^2 \left(1 + \sqrt{5}\frac{|x-y|}{l} + \frac{5}{3}\frac{|x-y|^2}{l^2}\right) \exp\left(- \sqrt{5} \frac{|x-y|}{l}\right),
\end{equation*}
with hyperparameters $\sigma$ and $\lambda$, which correspond to the standard deviation and the correlation length, respectively.

By representing $y(x)$ via the CKLE~\labelcref{eq:ckle_y}, we replace the discrete vector $\mathbf{y}$ as the unknown of the inverse problem with the CKLE coefficients $\boldsymbol{\xi}$.
Specifically, we propose parameterizing $\mathbf{y}$ in the MAP problem~\labelcref{eq:pde-constrained-opt-reg} via the discrete CKLE
\begin{equation}
  \label{eq:ckle_y_discrete}
  \mathbf{y}^c(\boldsymbol{\xi}) \equiv \bar{\mathbf{y}}^c + \boldsymbol{\Psi}_{\mathbf{y}} \boldsymbol{\xi},
\end{equation}
where
\begin{equation*}
  \left [ \bar{\mathbf{y}}^c \right]_i \equiv \bar{y}^c(\hat{x}_i), \quad \left [ \boldsymbol{\Psi}_{\mathbf{y}} \right ]_{ij} \equiv \sqrt{\lambda^y_j} \phi^y_j(\hat{x}_i).
\end{equation*}
We refer to this approach as the ``CKLEMAP'' method.
Given that, for sufficiently smooth log-transmissivity fields, the number of CKLE coefficients required to accurately represent $y^c$ is much smaller than the number of FV cells, i.e., $N_y \ll N$, the CKLEMAP method is less computationally expensive than the MAP method.

%

\subsection{CKLEMAP minimization problem formulation}

By solving~\cref{eq:pde-discretized} with $\mathbf{y} = \mathbf{y}^c(\boldsymbol{\xi})$, it can be seen that $\mathbf{u}$ can be expressed as a function of $\boldsymbol{\xi}$; specifically,
\begin{equation}
  \label{eq:inverse_problem}
  \mathbf{u}(\boldsymbol{\xi}) = \left[\mathbf{A}(\boldsymbol{\xi})\right]^{-1} \mathbf{b}(\boldsymbol{\xi}),
\end{equation}
where $\mathbf{A}\left(\boldsymbol{\xi}\right) = \mathbf{A}\left(\mathbf{y}^c(\boldsymbol{\xi})\right)$ and $\mathbf{b}\left(\boldsymbol{\xi}\right) = \mathbf{b}\left(\mathbf{y}^c(\boldsymbol{\xi})\right)$.
By expressing $\mathbf{u}$ as a function of $\boldsymbol{\xi}$, we can remove the PDE constraint from~\cref{eq:pde-constrained-opt-reg}, leading to the CKLEMAP unconstrained minimization problem
\begin{equation}
  \label{eq:cklemap}
  \min_{\boldsymbol{\xi}} \quad \frac{1}{2} \| \mathbf{u}_{\mathrm{s}} - \mathbf{H}_{\mathbf{u}} \mathbf{u}(\boldsymbol{\xi}) \|^2_2 + \frac{1}{2} \| \mathbf{y}_{\mathrm{s}} - \mathbf{H}_{\mathbf{y}} \mathbf{y}^c (\boldsymbol{\xi}) \|^2_2 + \frac{\gamma}{2} \| \mathbf{D} \mathbf{y}^c (\boldsymbol{\xi}) \|^2_2.
\end{equation}

To solve the CKLEMAP problem~\cref{eq:cklemap}, we recast it as the nonlinear least-squares minimization problem
\begin{equation*}
  \min_{\boldsymbol{\xi}} \quad \frac{1}{2} \left \| \mathbf{f}(\boldsymbol{\xi}) \right \|^2_2, \quad \mathbf{f}(\boldsymbol{\xi}) =
  \begin{bmatrix}
    \mathbf{u}_{\mathrm{s}} - \mathbf{H}_{\mathbf{u}} \mathbf{u}(\boldsymbol{\xi})\\
    \mathbf{y}_{\mathrm{s}} - \mathbf{H}_{\mathbf{y}} \mathbf{y}^c (\boldsymbol{\xi})\\
    \sqrt{\gamma} \, \mathbf{D} \mathbf{y}^c(\boldsymbol{\xi})
  \end{bmatrix},
\end{equation*}
which we solve using the Trust Region Reflective algorithm~\citep{branch-1999-JSC}.
The least-square minimization algorithm requires the evaluation of the Jacobian $\mathbf{J}_{\boldsymbol{\xi}}$ of the objective vector of the least-squares problem, $\mathbf{f}$, which is also the most computationally demanding part of the least-square minimization.
This Jacobian evaluation is done in two steps.
First, we evaluate the Jacobian of the objective vector with respect to $\mathbf{y}^c$, which reads
\begin{equation}
    \label{eq:jac_xi_q}
    \mathbf{J}_{\boldsymbol{\xi}} = \mathbf{J}_{\mathbf{y}^c} \begin{bmatrix}
      \frac{\partial \mathbf{y}^c}{\partial \boldsymbol{\xi}} \\ \mathbf{I}
    \end{bmatrix} =
  \begin{bmatrix}
    - \mathbf{H}_{\mathbf{u}} \frac{\partial \mathbf{u}(\mathbf{y}^c)}{\partial \mathbf{y}^c}\\
    -\mathbf{H}_{\mathbf{y}}\\
    \sqrt{\gamma} \, \mathbf{D}
  \end{bmatrix} \begin{bmatrix}
      \boldsymbol{\Psi}_{\mathbf{y}} \\ \mathbf{I}
    \end{bmatrix}.
\end{equation}
The partial derivative $\partial \mathbf{u} / \partial \mathbf{y}^c$ is evaluated via the chain rule~\citep{barajassolano-2014-linear,yeung-2022-physics} as described in Section~\ref{sec:jacobian}.
We note that most elements of $\mathbf{J}_{\mathbf{y}^c}$ are constant over iterations except the partial derivatives in the first block row.
These constant values are computed once before the least-square minimization and reused in each iteration.
With $\mathbf{J}_{\mathbf{y}^c}$ computed, $\mathbf{J}_{\boldsymbol{\xi}}$ can then be evaluated by postmultiplying the first block column by $\boldsymbol{\Psi}_{\mathbf{y}}$.

\subsection{Computations of partial derivatives in the evaluation of Jacobian}\label{sec:jacobian}

In this section we describe how the partial derivative $\partial \mathbf{u} / \partial \mathbf{y}^c$, required to evaluate the Jacobian of~\cref{eq:jac_xi_q}, are evalauted.
Let $p$ denote $y^c_i$.
Differentiating~\cref{eq:pde-discretized} with respect to $p$ yields
\begin{equation}
  \frac{\mathrm{d} \mathbf{l}}{\mathrm{d} p} = \frac{\partial \mathbf{l}}{\partial \mathbf{u}} \frac{\partial \mathbf{u}}{\partial p} + \frac{\partial \mathbf{l}}{\partial p} = \mathbf{A} \frac{\partial \mathbf{u}}{\partial p} + \left ( \frac{\partial \mathbf{A}}{\partial p} \mathbf{u} - \frac{\partial \mathbf{b}}{\partial p} \right ) = 0,
\end{equation}
which can be readily solved for $\partial \mathbf{u} / \partial p$, leading to the expression
\begin{equation}
  \label{eq:dudp}
  \frac{\partial \mathbf{u}}{\partial p} = -\mathbf{A}^{-1} \left ( \frac{\partial \mathbf{A}}{\partial p} \mathbf{u} - \frac{\partial \mathbf{b}}{\partial p} \right ) = -\mathbf{A}^{-1} \left. \frac{\partial \mathbf{l}}{\partial p} \right|_{\mathbf{u}}.
\end{equation}
It can be seen that evaluating $\partial \mathbf{u} / \partial \mathbf{y}^c$ requires evaluating the sensitivities of the TPFA stiffness matrix $\mathbf{A}$ and right-hand side vector $\mathbf{b}$ with respect to $\mathbf{y}^c$.
Substituting \cref{eq:dudp} into the first row block of \cref{eq:jac_xi_q} and taking the transpose yields
\begin{equation}
  \label{eq:jac_y_q_r1}
  \left [%
    \left. \dfrac{\partial \mathbf{l}}{\partial \mathbf{y}^c} \right|_{\mathbf{u}}
  \right ]^\top \mathbf{A}^{-1} \mathbf{H}^\top_{\mathbf{u}},
\end{equation}
by the fact that $\mathbf{A}$ is symmetric.

Note that in the MAP method, the Jacobian is given as
\begin{equation}
    \label{eq:jac_y_q}
    \mathbf{J}_{\mathbf{y}} = 
  \begin{bmatrix}
    - \mathbf{H}_{\mathbf{u}} \frac{\partial \mathbf{u}(\mathbf{y})}{\partial \mathbf{y}} \\
    -\mathbf{H}_{\mathbf{y}}\\
    \sqrt{\gamma} \, \mathbf{D}\\
  \end{bmatrix},
\end{equation}
and the partial derivatives are computed as in the CKLEMAP method, with $\mathbf{y}$ being treated the same way as $\mathbf{y}^c$. 

\subsection{Accelerated CKLEMAP method}

\begin{figure}
    \centering
    \begin{tikzpicture}
    \node[vertex] at (0.5, 2) (V1) {1};
    \node[vertex] at (0, 0) (V2) {2};
    \node[vertex, fill = midnight!20] at (2, 1) (V3) {3};
    \node[vertex, fill = midnight!20] at (3, 1.5) (V4) {4};
    \node[vertex] at (3.5, 0) (V5) {5};
    \node[vertex, fill = midnight!20] at (4.5, 0.5) (V6) {6};
    \node[vertex, fill = midnight!20] at (6, 1.5) (V7) {7};
    \node[vertex, fill = midnight!20] at (7, 0) (V8) {8};
    \path[arrow]
        (V1) edge (V2)
        (V1) edge (V3)
        (V2) edge (V3)
        (V1) edge (V4)
        (V3) edge (V4)
        (V3) edge (V6)
        (V5) edge (V6)
        (V4) edge (V7)
        (V6) edge (V8)
        (V7) edge (V8);
\end{tikzpicture}
    \caption{Closure of a unit column vector $\mathbf{e}_3 \equiv [0, 0, 1, 0, \ldots]^\top$ in a graph $G(A)$. The nonzero entries of $\mathbf{A}^{-1} \mathbf{e}_3$ are those nodes in the closure, i.e., $\{3, 4, 6, 7, 8\}$.}\label{fig:closure}
\end{figure}
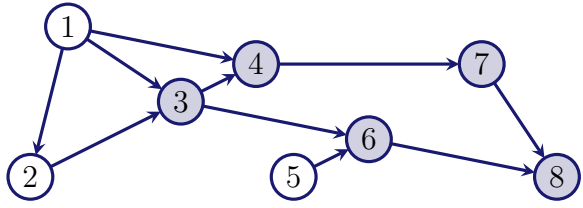

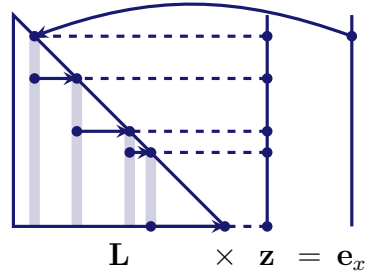
\begin{figure}
    \begin{subfigure}[b]{.55\textwidth}
        \begin{algorithmic}[1]
            \Procedure {FindSparsity}{$\mathbf{L}$, $x$}
            \State $j\gets x$
            \State $\mathbb{S}\gets \{j\}$
            \While{$j\neq N$}
                \State $j\gets \mathrm{argmin}_{i > j} \mathbf{L}[i, j] \neq 0$
                \State $\mathbb{S}\gets \mathbb{S} \cup \{j\}$
            \EndWhile
            \State \Return $\mathbb{S}$
            \EndProcedure
        \end{algorithmic}
        \caption{Algorithm}\label{fig:pseudocode}
    \end{subfigure}
    \begin{subfigure}[b]{.4\textwidth}
        \centering
        \begin{tikzpicture}[x = 4pt, y = 4pt]
            \fill[fill = midnight!20] (1.5, 0) rectangle (2.5, 18)
                (5.5, 0) rectangle (6.5, 14)
                (10.5, 0) rectangle (11.5, 9)
                (12.5, 0) rectangle (13.5, 7);
            \draw (0, 0) -- (0, 20) -- (20, 0) -- cycle
                (24, 0) -- (24, 20)
                (32, 0) -- (32, 20);
            \fill (32, 18) coordinate (P1) circle (0.5)
                (2, 18) coordinate (P2) circle (0.5)
                (2, 14) coordinate (P3) circle (0.5)
                (6, 9) coordinate (P4) circle (0.5)
                (11, 7) coordinate (P5) circle (0.5)
                (13, 0) coordinate (P6) circle (0.5)
                (6, 14) coordinate (P7) circle (0.5)
                (11, 9) coordinate (P8) circle (0.5)
                (13, 7) coordinate (P9) circle (0.5)
                (20, 0) coordinate (P10) circle (0.5)
                (24, 18) coordinate (P11) circle (0.5)
                (24, 14) coordinate (P12) circle (0.5)
                (24, 9) coordinate (P13) circle (0.5)
                (24, 7) coordinate (P14) circle (0.5)
                (24, 0) coordinate (P15) circle (0.5);
            \draw[arrow] (P1) [out = 160, in = 20] to (P2);
            \draw[arrow] (P3) -- (P7);
            \draw[arrow] (P4) -- (P8);
            \draw[arrow] (P5) -- (P9);
            \draw[arrow] (P6) -- (P10);
            \draw[dashed] (P11) -- (P2)
                (P12) -- (P7)
                (P13) -- (P8)
                (P14) -- (P9)
                (P15) -- (P10);
            \node[var] at (10, -10pt) (A) {$\mathbf{L}$};
            \node[var] at (20, -10pt) {$\times$};
            \node[var] at (24, -10pt) (x) {$\mathbf{z}$};
            \node[var] at (28, -10pt) {$=$};
            \node[var] at (32, -10pt) (b) {$\mathbf{e}_x$};
        \end{tikzpicture}
        \caption{Graphical Example}\label{fig:graphical_example}
    \end{subfigure}
    \caption{Algorithm for finding the sparsity structure $\mathbb{S}$ of $\mathbf{z} = \mathbf{L}^{-1} \mathbf{e}_x$.}\label{fig:partial_solve}
\end{figure}

\begin{figure}
    \centering
    \includegraphics{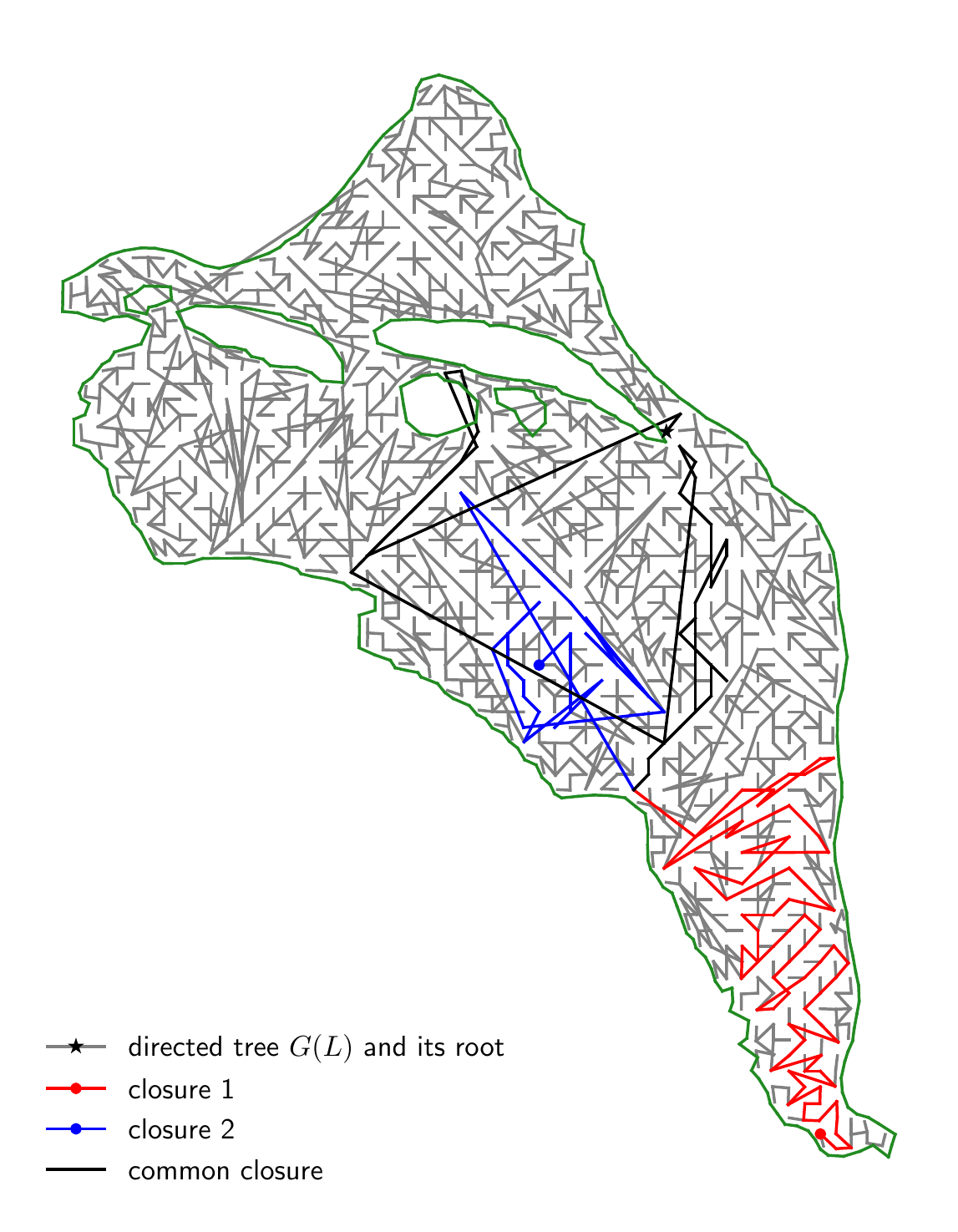}
    \caption{The directed tree $G(L)$ structure on with two closures from different cells.}\label{fig:hanford_closure}
\end{figure}

In the ``accelerated'' CKLEMAP method, we compute $\mathbf{A}^{-1} \mathbf{H}^\top_{\mathbf{u}}$ efficiently by exploiting the sparsity structure of the Cholesky factor of $\mathbf{A}$.
Recall that each column of $\mathbf{H}^\top_{\mathbf{u}} = (\mathbf{I}_N [\mathcal{I}_u])^{\top}$ has only one non-zero entry.
Therefore, if the sparsity structure of the Cholesky factor $\mathbf{L}$ of $\mathbf{A}$ is known, the sparsity structure of each column of $\mathbf{Z} = \mathbf{L}^{-1} \mathbf{H}^\top_{\mathbf{u}}$ is $\{\mathrm{closure}_{\mathbf{L}}(i) \mid i \in \mathcal{I}_u\}$, that is, the subset of vertices in the graph $G(\mathbf{L})$ that have a path from each vertex $i \in \mathcal{I}_u$~\citep{gilbert-1994-JMAA}.
Figure~\ref{fig:closure} shows an example of a closure.
Furthermore, the graph of a Cholesky factor $\mathbf{L}$ is a directed tree, and any closure induced by a vertex $i$ is all the vertices along the path from $i$ to the root of the tree~\citep{yeung-2016-toc}.
This enables a simple algorithm to find the sparsity structure of the solution of $\mathbf{L} \mathbf{Z} = \mathbf{H}^{\top}_{\mathbf{u}}$.
Figure~\ref{fig:partial_solve} illustrates this algorithm together with a graphical example.
Once we have the sparsity structure $\mathcal{Z}_i$ of $\mathbf{z}_i$, the column $i$ of $\mathbf{Z}$, we only need the submatrix $\mathbf{L}[\mathcal{Z}_i, \mathcal{Z}_i]$ instead of the whole matrix $\mathbf{L}$ to solve for $\mathbf{z}_i$. Such submatrix is highlighted in blue dots in the lower triangular matrix $\mathbf{L}$ in Figure~\ref{fig:graphical_example}. This eliminates the unnecessary computations involving the part of $\mathbf{L}$ that does not contribute to the final solutions, thus accelerating the computations.
Furthermore, since the topology of the FV discretization is static, the sparsity structure of the Cholesky factor $\mathbf{L}$ is fixed throughout the entire least-square minimization procedure.
Given this, together with the  fact that $\mathbf{H}_{\mathbf{u}}$ is constant, it follows that $\mathcal{Z}_i$ is also fixed and only needs to be computed once.
Figure~\ref{fig:hanford_closure} shows the closures of two observation locations in $\mathcal{I}_u$ on the Hanford Site experiment to be discussed in detail in~\cref{sec:experiments}. The gray lines indicate the cells that do not contribute to the columns of the Jacobian corresponding to either of these two locations.

We note that, although the computations of the Jacobian can be accelerated by 3--4 times using the procedure described above, the overall execution time reduction in solving the minimization problems exhibited by the numerical experiments of Section~\ref{sec:experiments} is 10--20\%.
This is because the nonlinear least-squares minimization algorithm, the Trust Region algorithm, dominates most of the execution time.
The execution times can be further reduced by optimizing the implementation of the Trust Region algorithm.

\section{Numerical experiments}\label{sec:experiments}

\subsection{Case study}

\begin{figure}
    \centering
    \includegraphics{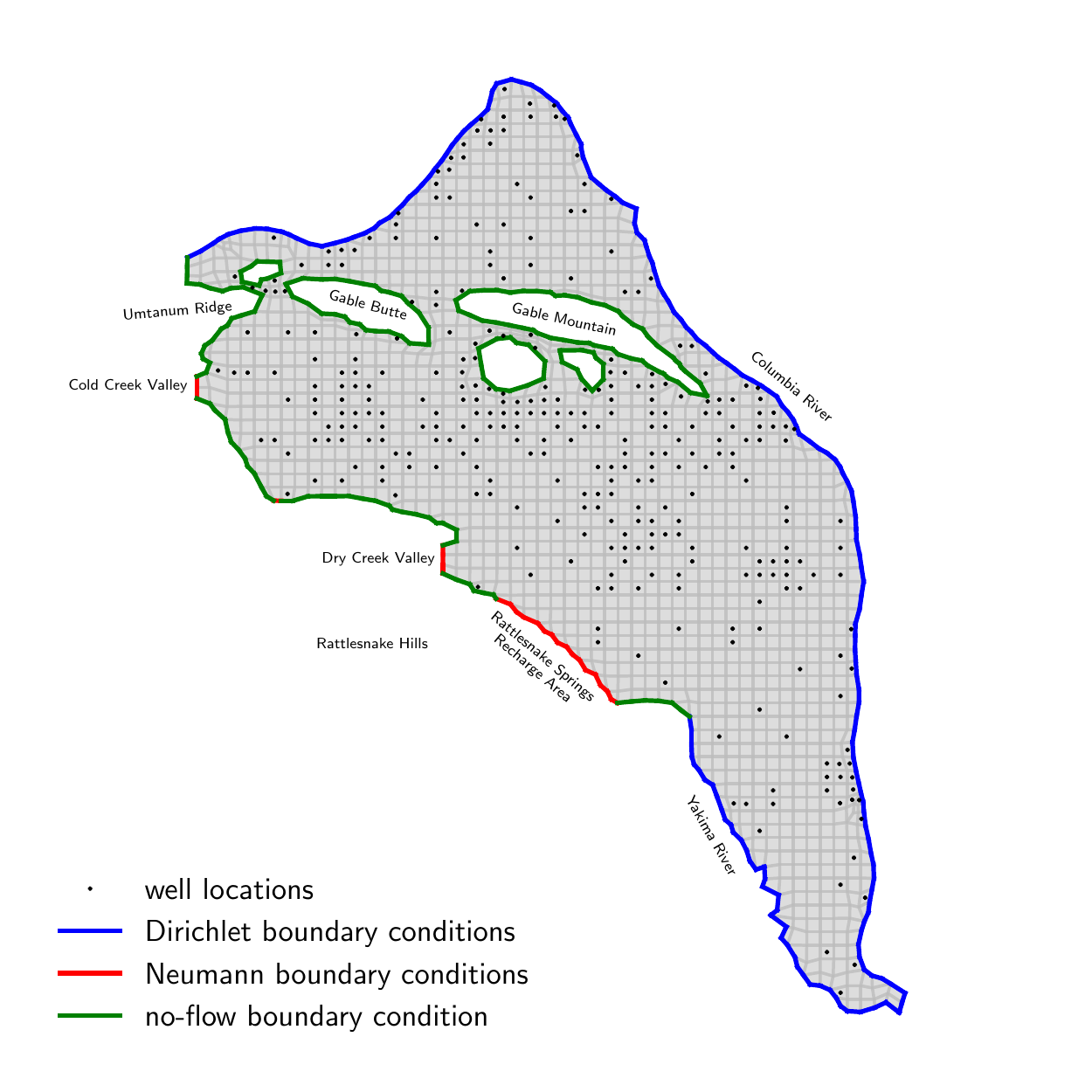}
    \caption{The coarse-resolution mesh of ($N_{FV}=1475$) cells with well locations marked, and the parts of boundaries colored for different types of prescribed boundary conditions.}\label{fig:hanford_site}
\end{figure}

We evaluate the performance of the proposed CKLEMAP formulation against MAP with a case study of parameter estimation in a steady-state two-dimensional groundwater model of the Hanford Site. 
The reference log-transmissivity field $\tilde{y}$ and boundary conditions $u_\mathcal{D}$ and $q_\mathcal{N}$ are based on the data obtained from a three-dimensional Hanford Site calibration study~\citep{cole-2001-transient} and are shown in Figure~\ref{fig:hanford_site}.
The details of the reference transmissivity field generation are given in~\cite{yeung-2022-physics}.
To study the scalability of the CKLEMAP and MAP methods with the problem size (i.e., the number of cells in the FV model), we generate the reference field at two additional resolutions with four times and 16 times the number of cells in the base FV model, respectively.

The numbers of cells in the low, medium, and high-resolution models are 1475, 5900, and 23600, respectively. 
For a higher resolution mesh, we divide each cell in a lower resolution model into four equiareal subcells and interpolate $\tilde{y}$ at the centers of each subcell, as well as $u_\mathcal{D}$ and $q_\mathcal{N}$ at the midpoints of each boundary edge of the boundary subcells.

There are 558 wells at the Hanford Site where $u$ can be potentially measured~\citep{cole-2001-transient}. Some of these wells are located in the same coarse or fine cells. Figure~\ref{fig:hanford_site} shows the locations of the cells in the low-resolution FV model that contain at least one well.
Since our model uses exclusively cells but not points to specify spatial locations, multiple wells are treated as a single well if they are located in the same cell.
As a result, there are 323 wells in the low-resolution FV model, while the medium-resolution model has 408 wells.

The aforementioned Hanford Site calibration study defined the Dirichlet and Neumann boundaries $\Gamma_\mathcal{D}$ and $\Gamma_\mathcal{N}$ as shown in Figure~\ref{fig:hanford_site}, and provides the estimates of the heads $u_\mathcal{D}$ and the fluxes $q_\mathcal{N}$ at these boundaries.
In setting boundary conditions for our comparison study, we assume that $u_\mathcal{D}$ and $q_\mathcal{N}$ are both known and are given by the estimate.

For each reference log-transmissivity field $\Tilde{y}$, we generate the hydraulic head field $\Tilde{u}$ by solving the Darcy flow equation on the corresponding FV mesh with the Dirichlet and (deterministic) Neumann boundary conditions that are set as described above.
The values of the reference $y$ and $u$ fields at all cell locations $\hat{x}_i$ are organized into the vectors $\Tilde{\mathbf{y}}$ and $\Tilde{\mathbf{u}}$, respectively.
Then, we randomly pick $N_{\mathbf{y}_{\mathrm{s}}}$ well locations and treat the values of $\Tilde{y}$ at these locations as $y$ measurements to form $\mathbf{y}_{\mathrm{s}}$.
Similarly, we draw $N_{\mathbf{u}_{\mathrm{s}}}$ measurements of the hydraulic head $u$ from $\Tilde{u}$ to form $\mathbf{u}_{\mathrm{s}}$.
These measurements are treated as synthetic data sets and used in the CKLEMAP and MAP methods to estimate the entire $y$ and $u$ fields.   

We note that the aquifer at the Hanford Site is unconfined, and the use of~\cref{eq:pde} to describe flow at the Hanford Site relies on a conceptual simplification.
A more accurate linear conceptual model for flow in an unconfined aquifer with a horizontal confining layer can be obtained based on the Dupuit–Forchheimer approximation in the form~\cite{zhang-2014-nonlinear}
\begin{align} \label{eq:Dupuit_pde}
  \nabla \cdot \left [ K(x) \nabla v(x) \right ]  = 0,  x \in D,
\end{align}
where $v(x) = u^2(x)$ and $K(x)$ is the depth-averaged conductivity.
Mathematically,~\cref{eq:Dupuit_pde,eq:pde} are identical, although the field $u(x)$ computed using these two equations will be different.
Therefore, solving the inverse problem for~\cref{eq:pde} is equivalent in complexity to solving the inverse problem for~\cref{eq:Dupuit_pde}.
We also note that applying the Dupuit–Forchheimer approximation to the Hanford Site aquifer will produce additional linear terms in~\cref{eq:Dupuit_pde} due to the variations in the elevation of the bottom confining layer of the aquifer.     

The implementation of CKLEMAP and MAP are written in Python using the NumPy and SciPy packages.
All CKLEMAP and MAP simulations are performed using a 3.2~GHz 8-core Intel Xeon W CPU and 32~GB of 2666~MHz DDR4 RAM\@.

The weight $\gamma$ in the CKLEMAP and MAP minimization problems is empirically found to minimize the error with respect to the reference $y$ fields as $\gamma=10^{-6}$. When a reference field is not known, these weights can be found using cross-validation~\citep{picard-1984-cross}.     

\subsection{Performance of CKLEMAP as a function of the number of KL terms}
\begin{table}[!htbp]
    \centering\tabulinesep=0.1em
    \caption{Performance of CKLEMAP in estimating the coarse-resolution (${N_{FV}=1475}$) mesh with $N_{\mathbf{y}_{\mathrm{s}}} = 100$ as functions of number of KL terms $N_y$.}\label{tab:kl_results}%
    \begin{tabu} to \textwidth {r*{5}{X[cm]}}
        \toprule
        & \multicolumn{5}{c}{$N_y$} \\
        & 200 & 400 & 600 & 800 & 1000 \\
        \midrule
        \makecell[r]{least square\\iterations} & 99--218 & 44--335 & 25--69 & 28--177 & 20--65 \\
        \midrule
        \makecell[r]{execution\\time (s)} & 17.55--42.14 & 12.37--86.31 & 9.76--24.73 & 14.60--94.86 & 14.25--36.29 \\
        \midrule
        \makecell[r]{relative\\$\ell_2$ error} & 0.265--0.568 & 0.137--0.239 & 0.081--0.098 & 0.072--0.082 & 0.072--0.083 \\
        \midrule
        \makecell[r]{absolute\\$\ell_\infty$ error} & 13.08--42.69 & 6.56--16.32 & 3.71--5.63 & 3.68--5.22 & 3.46--5.31 \\
        \bottomrule
    \end{tabu}
\end{table}

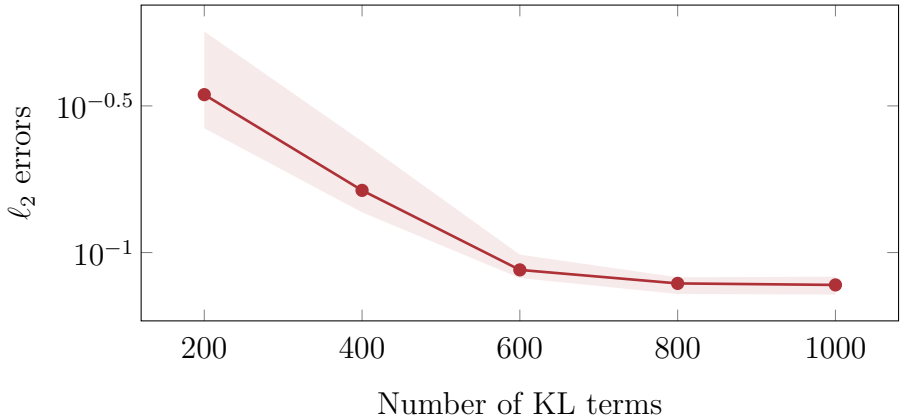
\begin{figure}[!htbp]
    \centering
    \begin{tikzpicture}
		\begin{semilogyaxis}[error plot]
			\errorband[Maroon,
			    every mark/.append style={solid,fill=Maroon},
				mark=*]
				{figures/rel_errors_vs_kl.txt}{kl}{rel_errors_min}{rel_errors_max}{rel_errors_avg};
		\end{semilogyaxis}
		
	\end{tikzpicture}
	\caption{Relative $\ell_2$ errors versus the number of KL terms.}\label{fig:rel_errors_vs_kl}
\end{figure}

First, we study the relative $\ell_2$ and absolute $\ell_\infty$ errors in the CKLEMAP solution for $y$ as well as the time-to-solution and the number of iterations of the minimization algorithm as functions of $N_y$, the number of terms in the CKLE of $y$ for $N_{\mathbf{y}_{\mathrm{s}}} = 100$. 
The relative $\ell_2$ and absolute $\ell_\infty$  errors are computed on the FV mesh, respectively, as
\begin{equation}
    \varepsilon_2 (y) \equiv \frac{ \|\hat{\mathbf{y}} - \Tilde{\mathbf{y}}\|_2 }{ \|\Tilde{\mathbf{y}}\|_2 }.
\end{equation}
and
\begin{equation}
    \varepsilon_\infty (y) \equiv \|\hat{\mathbf{y}} - \Tilde{\mathbf{y}}\|_\infty.
\end{equation}

We find that for the considered inverse problem, all these quantities strongly depend on the locations of $y$ measurements.
Therefore, we compute these quantities for 10 different distributions of the measurement locations.
The ranges of the $\ell_2$ and $\ell_\infty$ errors, execution times, and the numbers of iterations are reported in Table~\ref{tab:kl_results}. The $\ell_2$ error and its bounds as functions of $N_y$ are also plotted in Figure~\ref{fig:rel_errors_vs_kl}.
We find that the $\ell_2$ errors decrease with increasing $N_y$ and converge to asymptotic values for $N_y \approx 800$.
The lower bound of  $\ell_\infty$ continues to decrease even for $N_y$ greater than 800, while the upper bound increases from 5.22 to 5.31 as $N_y$ increases from 800 to 1000.
However, the relative changes of $\ell_\infty$ are insignificant for $N_y>800$.
What is surprising is that the execution time does not significantly change with increasing $N_y$.
While the time per iteration increases with $N_y$, the number of iterations tends to decrease.
Therefore, in the rest of the numerical examples, we set $Ny=1000$, which corresponds to $\text{rtol}_y$ on the order of $10^{-8}$.

\subsection{CKLEMAP and MAP errors versus the number of \texorpdfstring{$y$}{y} measurements}

Next, we study the accuracy of the CKLEMAP and MAP methods in estimating $y$ as the function of the number of $y$ measurements. We assume that $u$ measurements are available at all wells.
 
\begin{figure}[!htbp]
    \centering\tabulinesep=0.1em
    \begin{tabu} to \textwidth {X[1.25,cm]*{4}{X[cm]}l}
        reference & \multicolumn{4}{>{\centering\arraybackslash}m{0.65\textwidth}}{\includegraphics[scale=0.25]{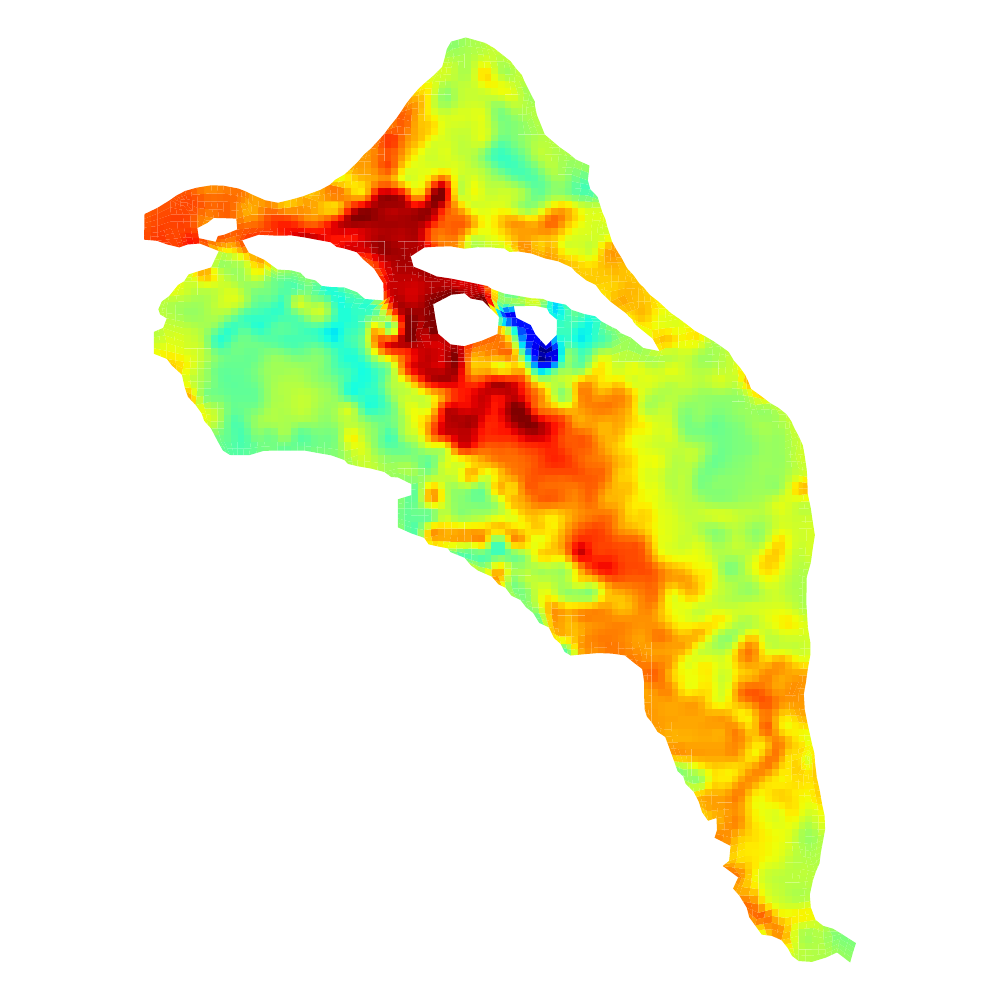}} &
        \colorbar{0}{12}{{0,2,4,...,12}} \\
        \midrule
        $N_{\mathbf{y}_{\mathrm{s}}}$ & 25 & 50 & 100 & 200 & \\
        \shortstack[c]{observation\\locations} &
        \includegraphics[scale=0.25]{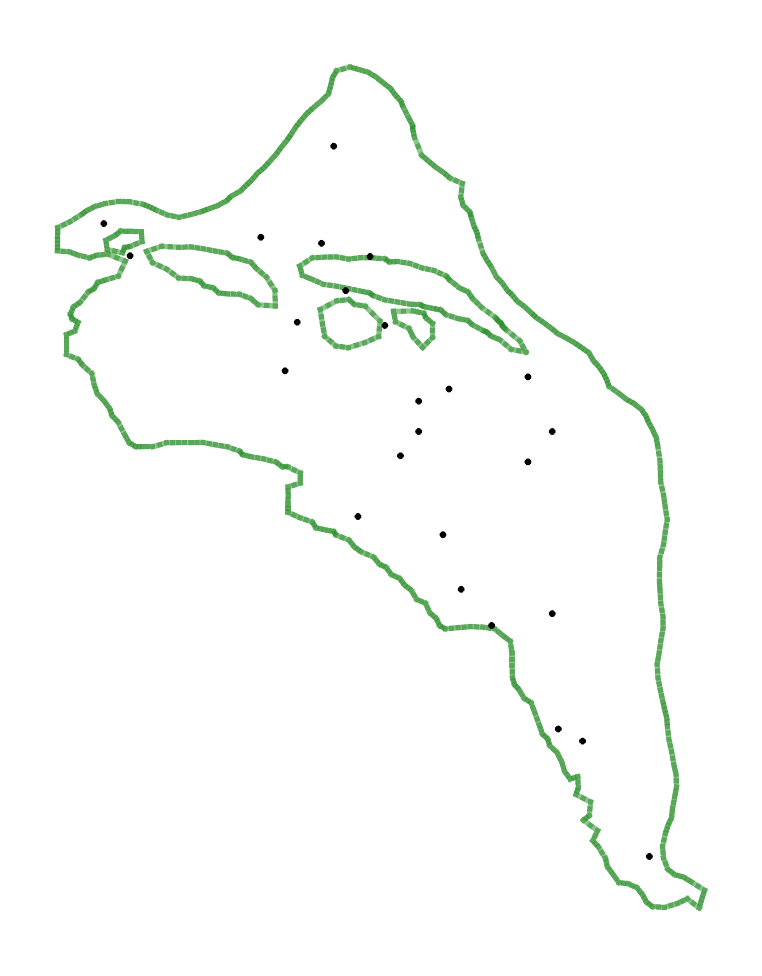} &
        \includegraphics[scale=0.25]{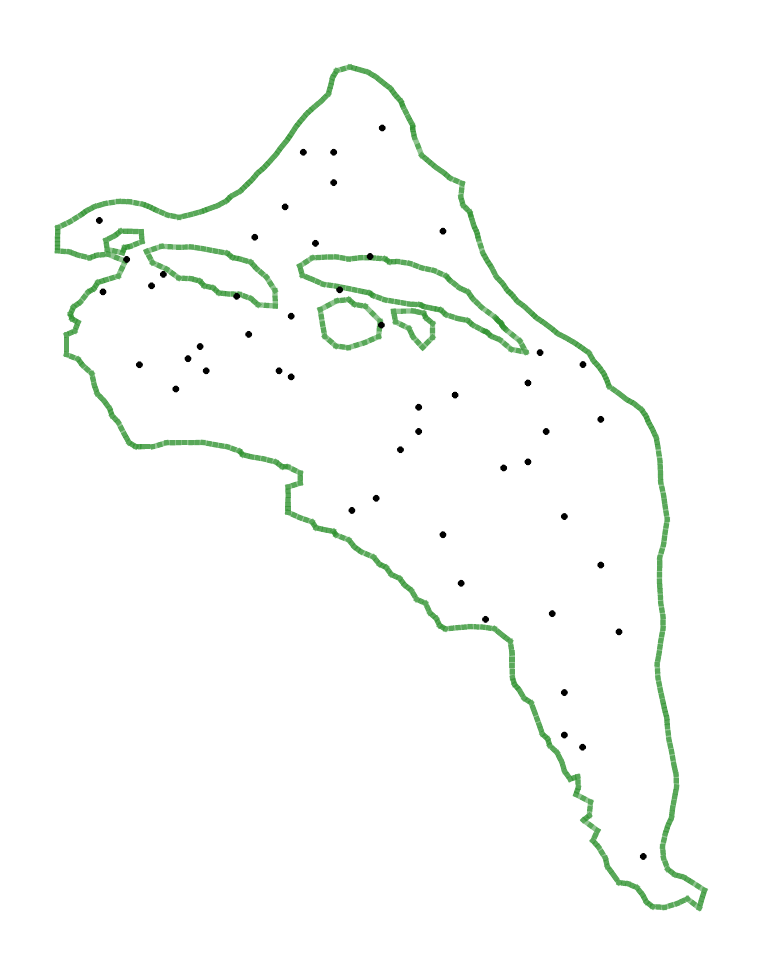} &
        \includegraphics[scale=0.25]{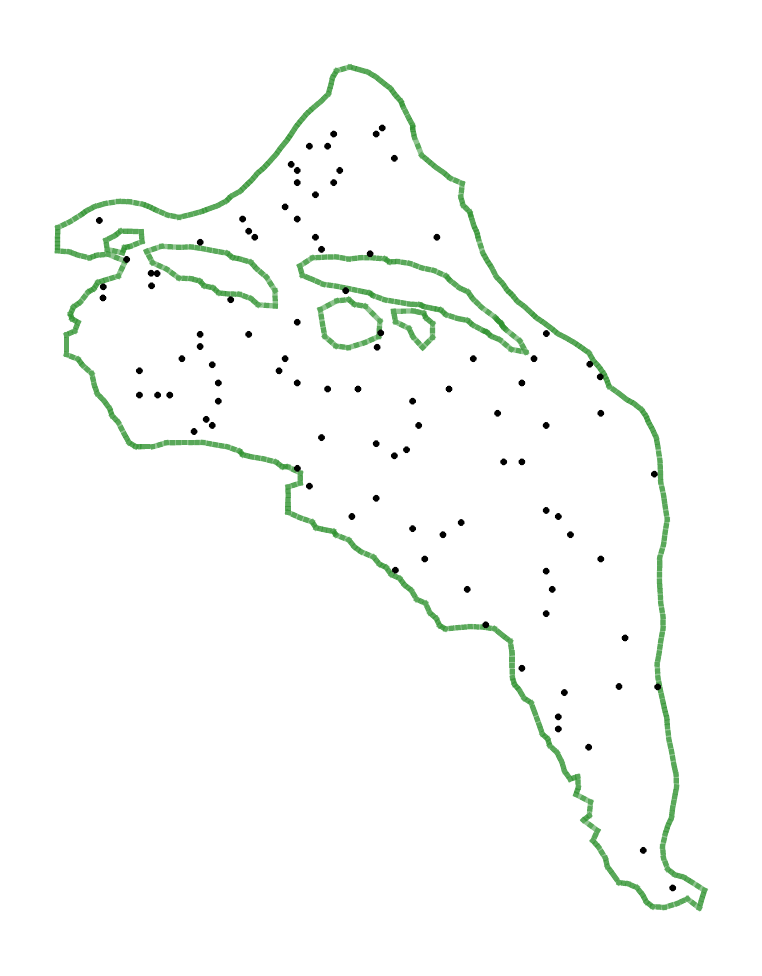} &
        \includegraphics[scale=0.25]{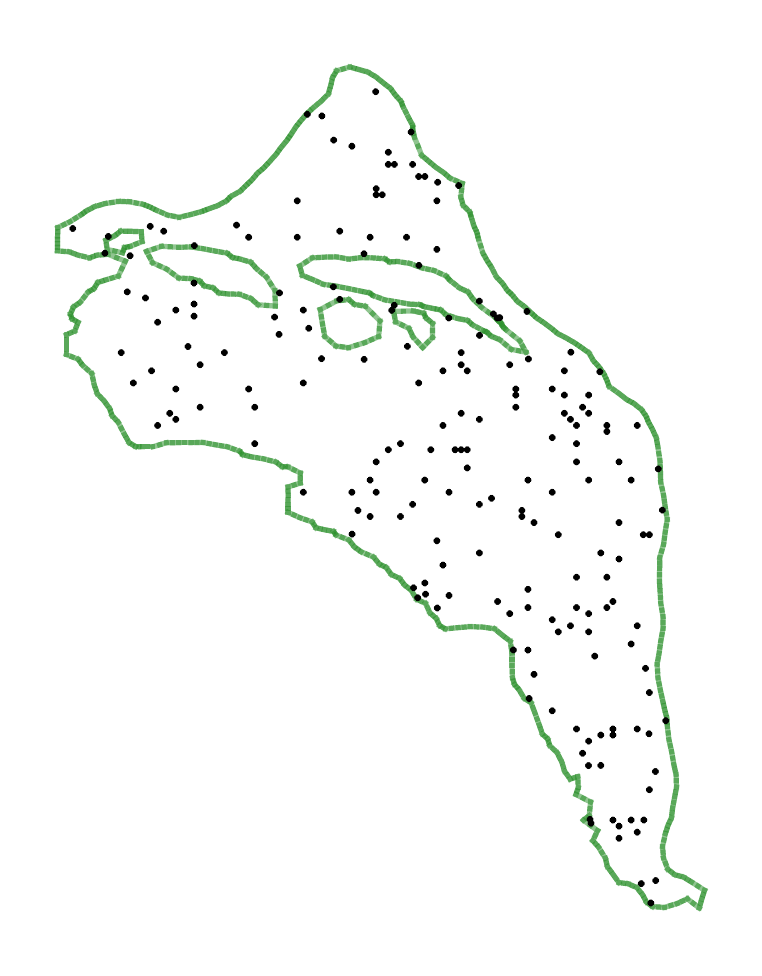} & \\
        \shortstack[c]{CKLEMAP\\estimates} &
        \includegraphics[scale=0.25]{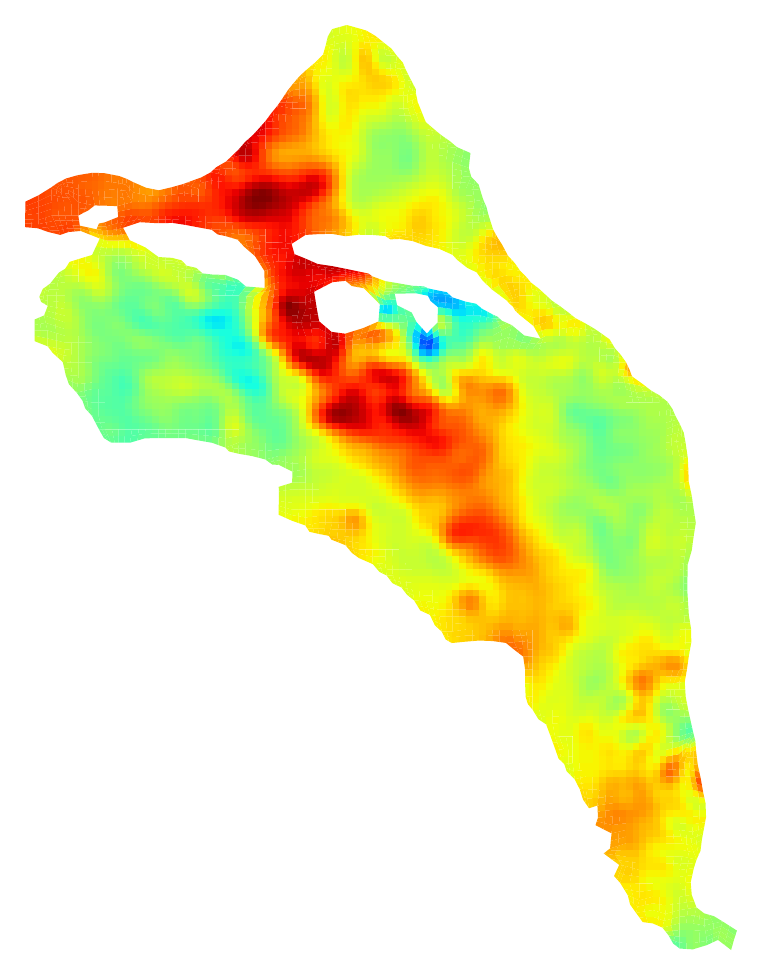} &
        \includegraphics[scale=0.25]{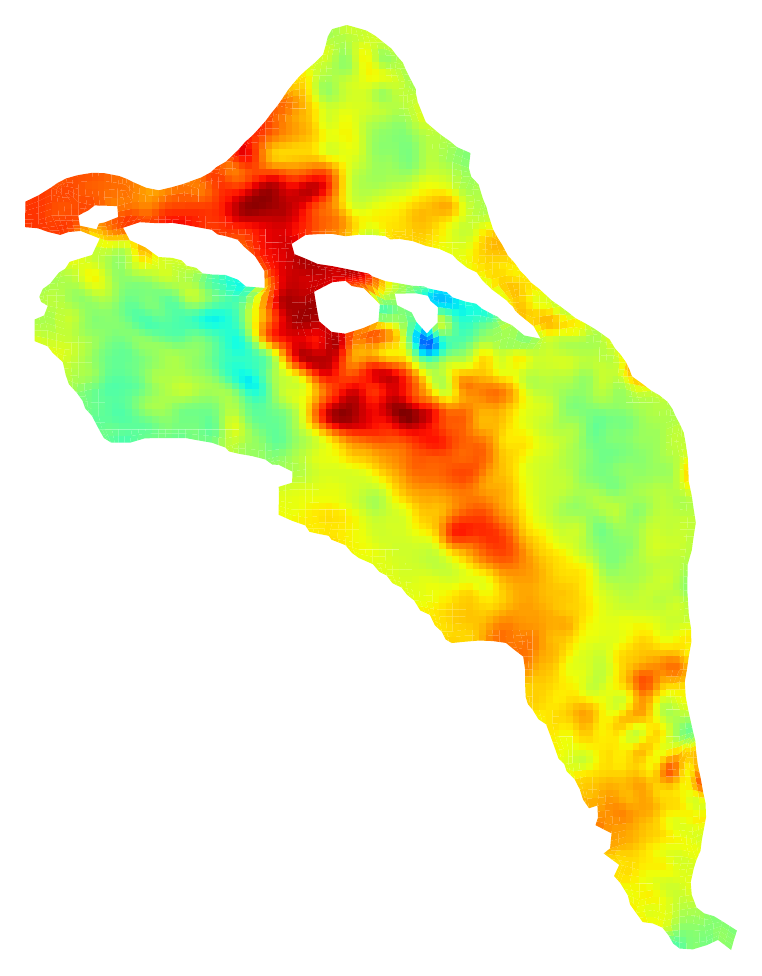} &
        \includegraphics[scale=0.25]{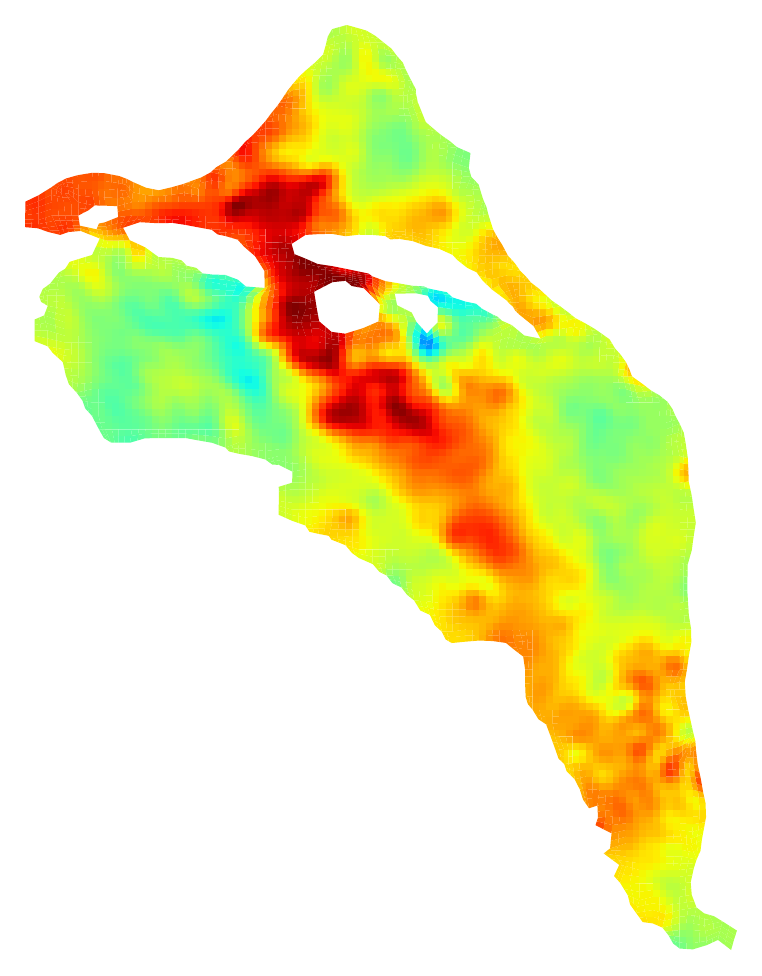} &
        \includegraphics[scale=0.25]{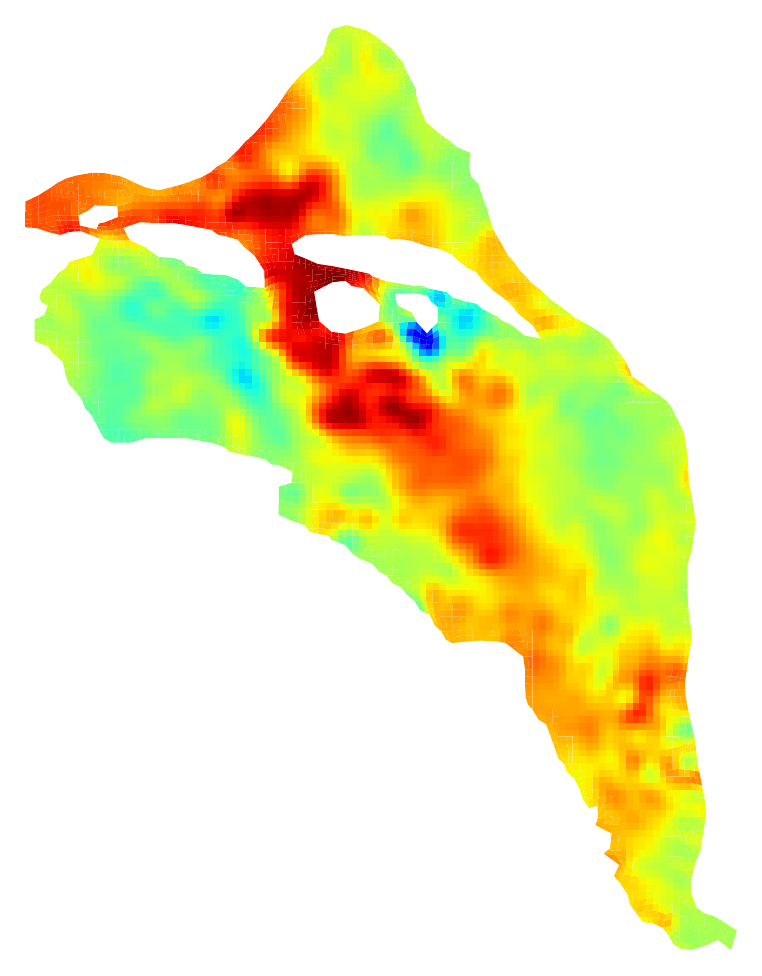} &
        \colorbar{0}{12}{{0,2,4,...,12}} \\
        \shortstack[c]{CKLEMAP\\point errors} & \includegraphics[scale=0.25]{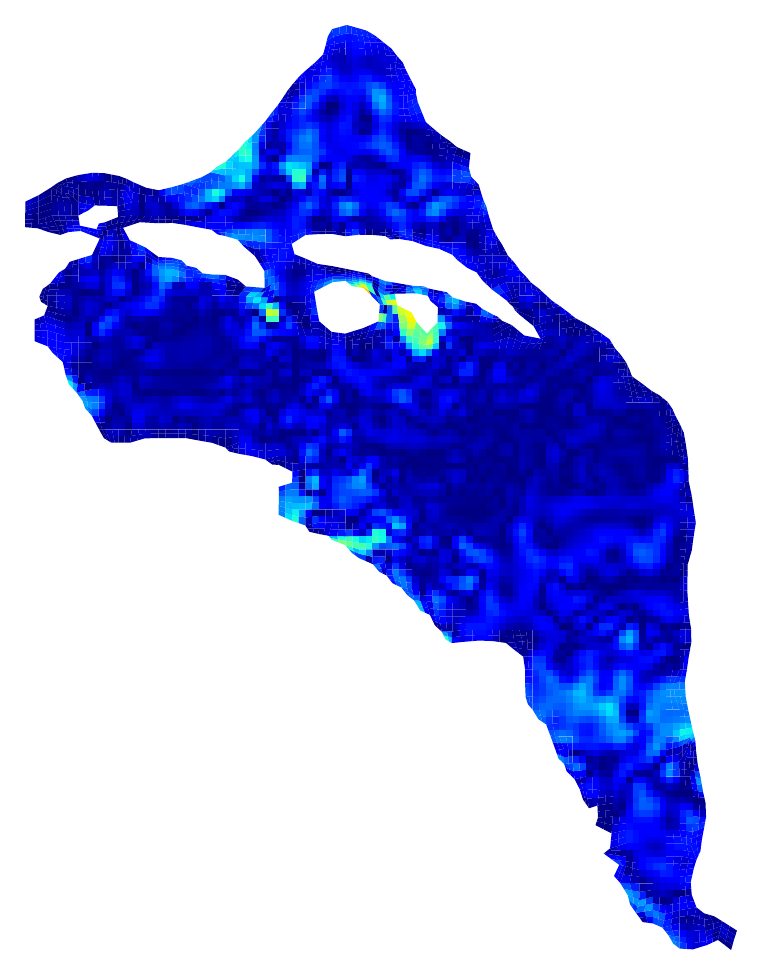} &
        \includegraphics[scale=0.25]{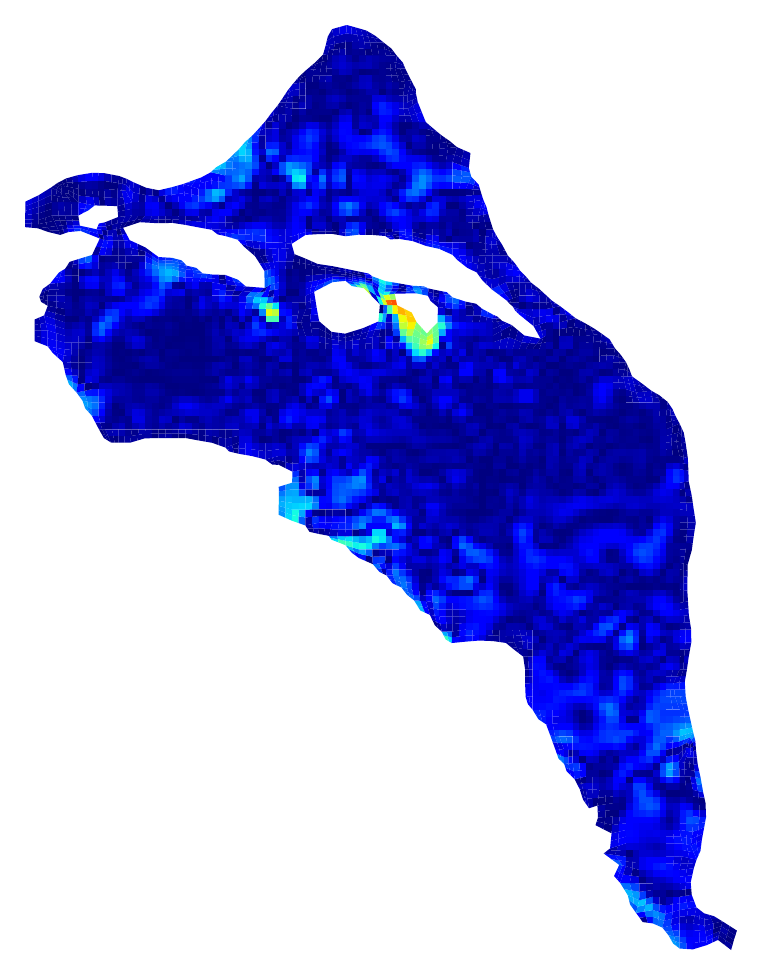} &
        \includegraphics[scale=0.25]{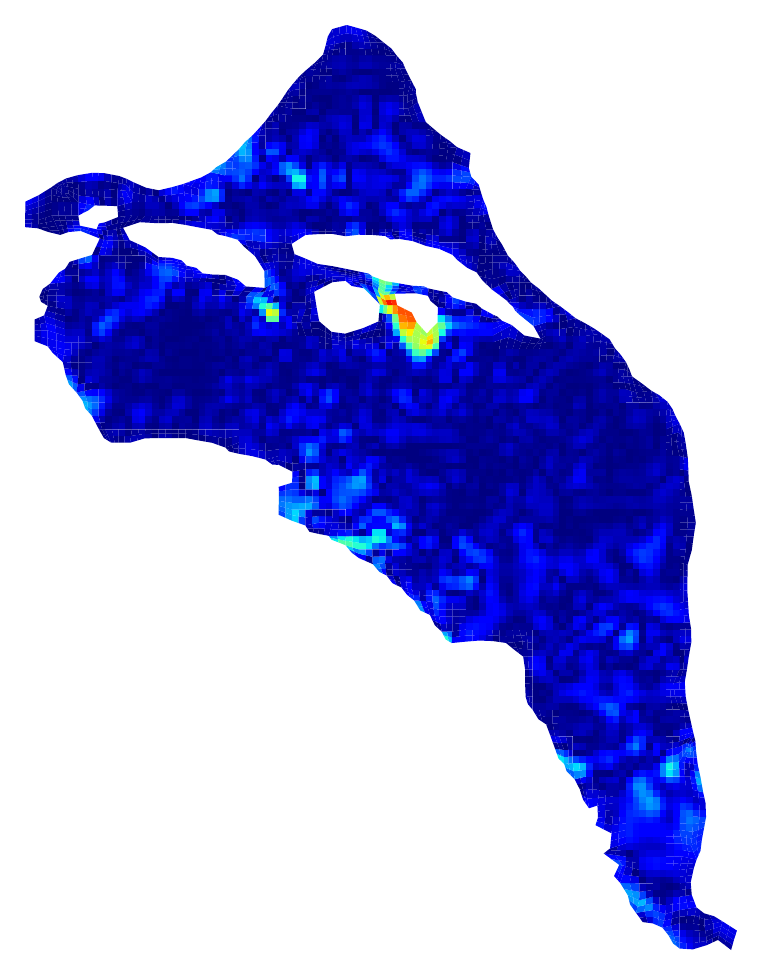} &
        \includegraphics[scale=0.25]{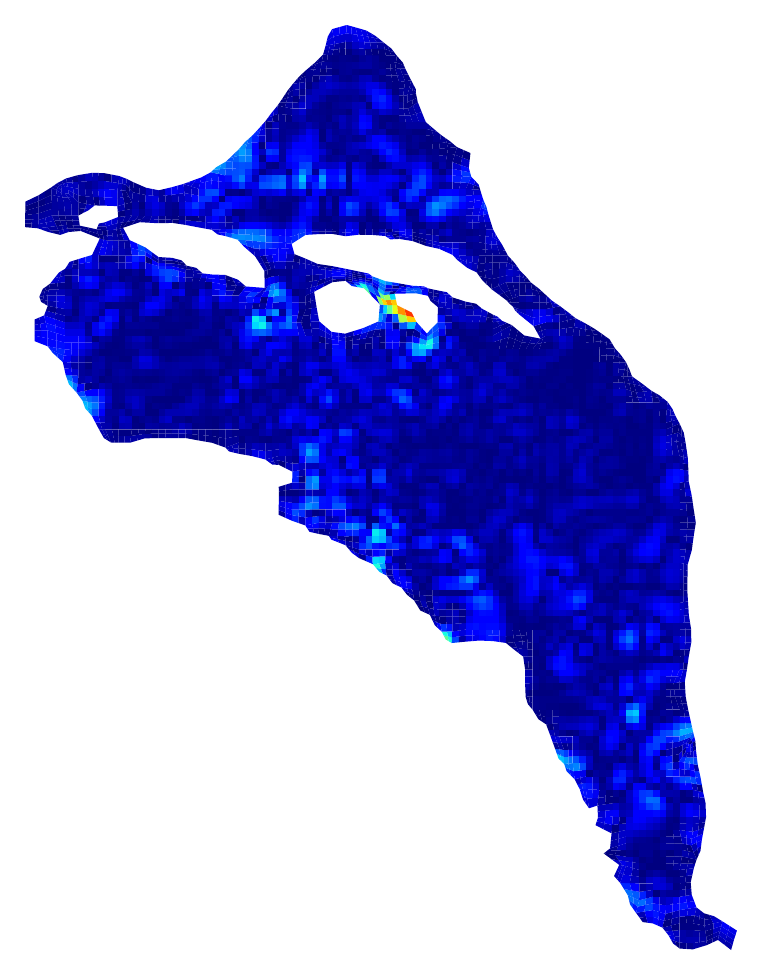} &
        \colorbar{0}{6}{{0,1,2,...,6}} \\
        \shortstack[c]{MAP\\estimates} &
        \includegraphics[scale=0.25]{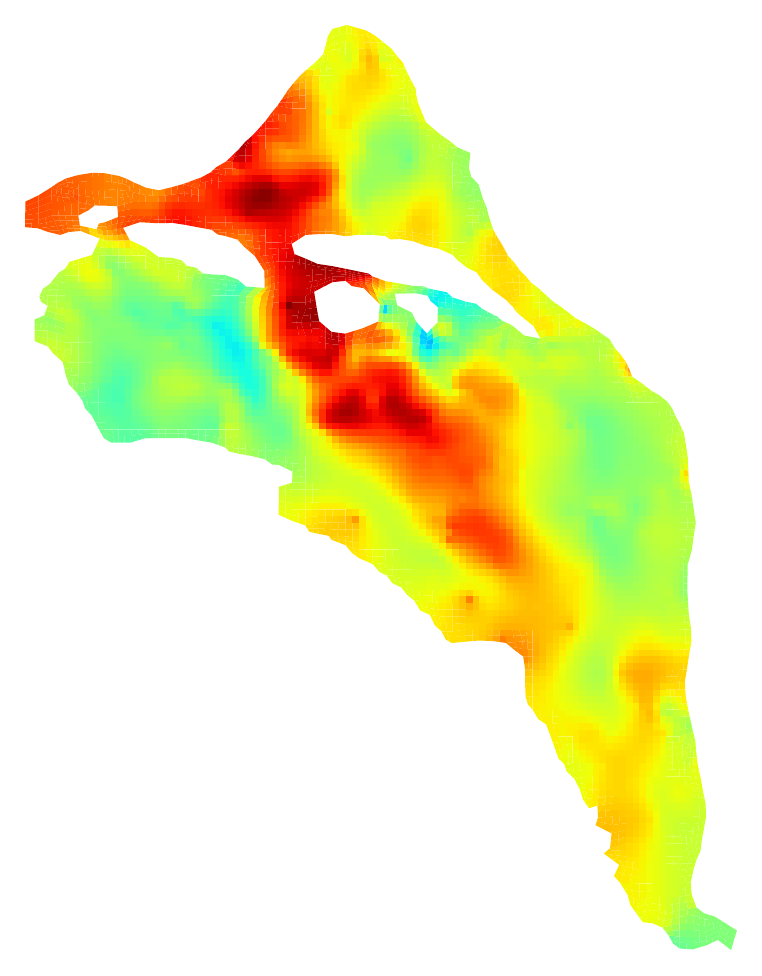} &
        \includegraphics[scale=0.25]{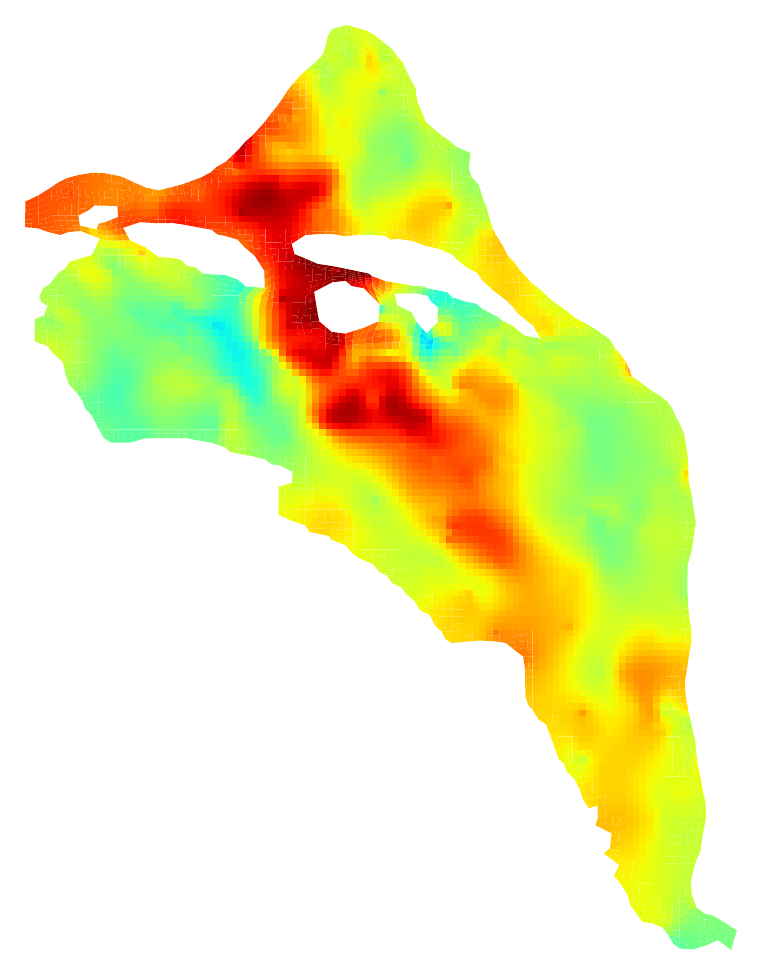} &
        \includegraphics[scale=0.25]{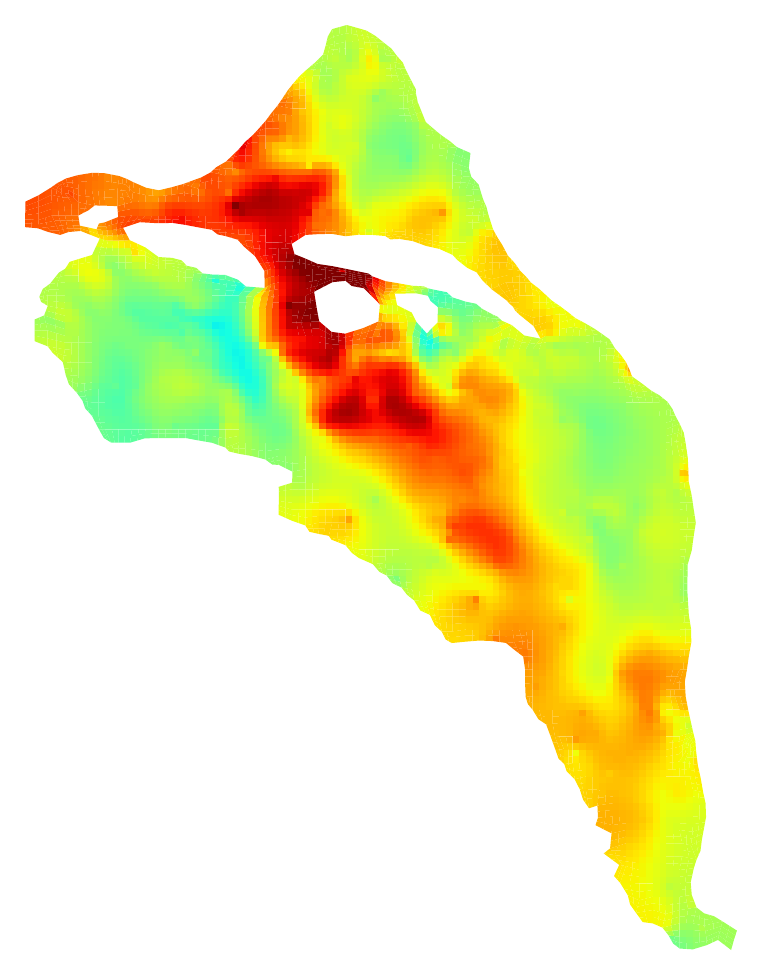} &
        \includegraphics[scale=0.25]{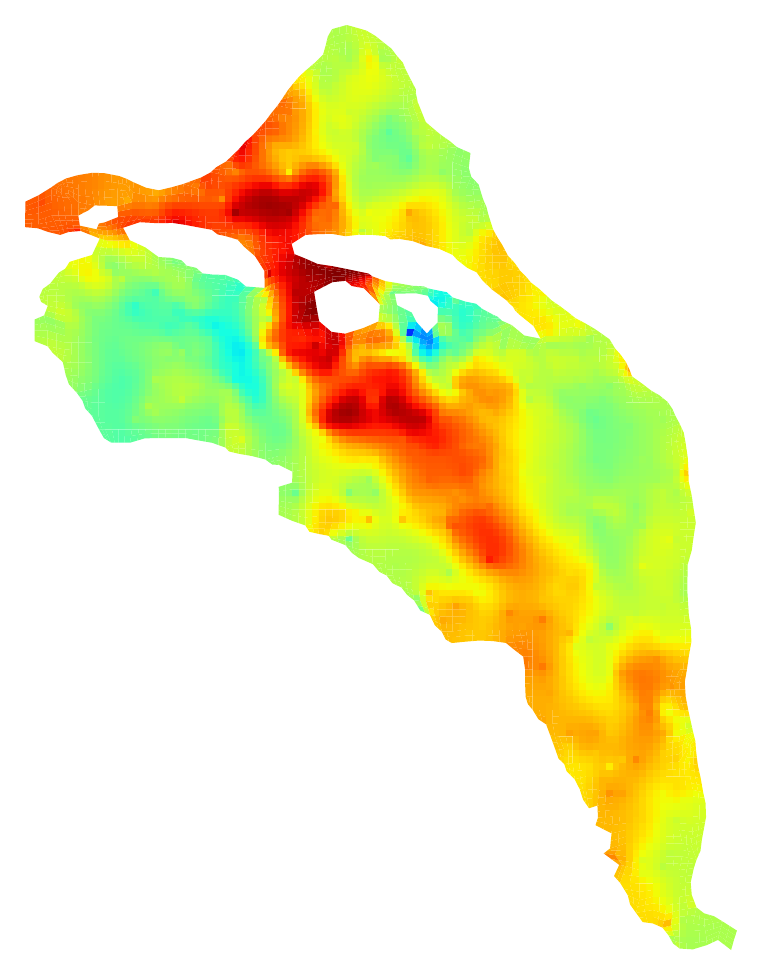} &
        \colorbar{0}{12}{{0,2,4,...,12}} \\
        \shortstack[c]{MAP\\point errors} & \includegraphics[scale=0.25]{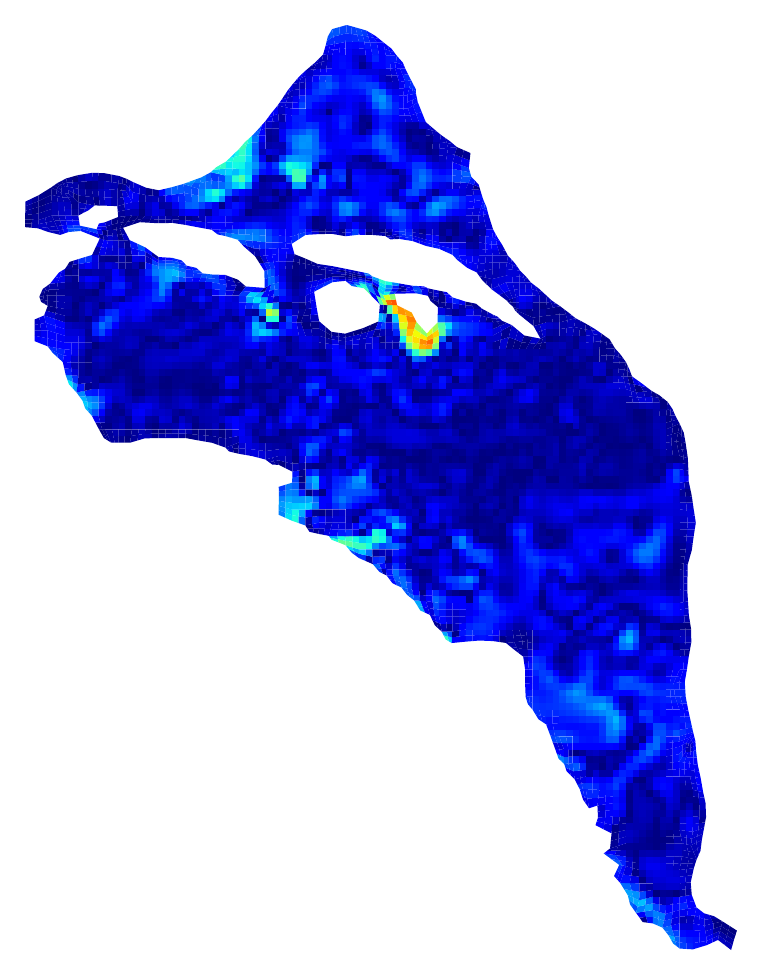} &
        \includegraphics[scale=0.25]{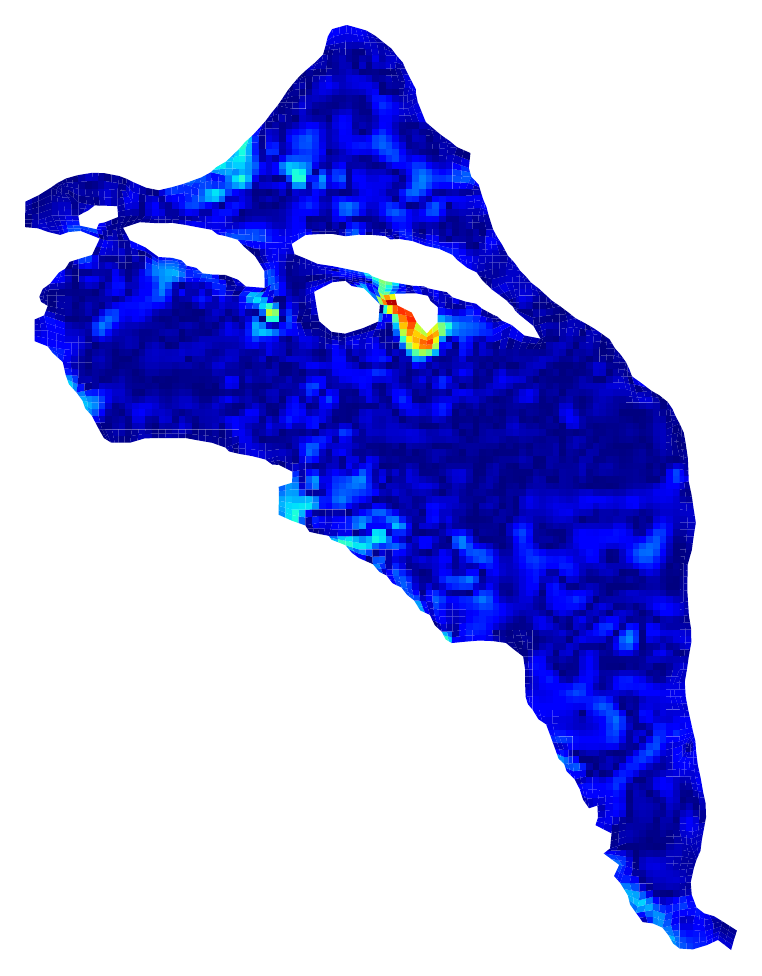} &
        \includegraphics[scale=0.25]{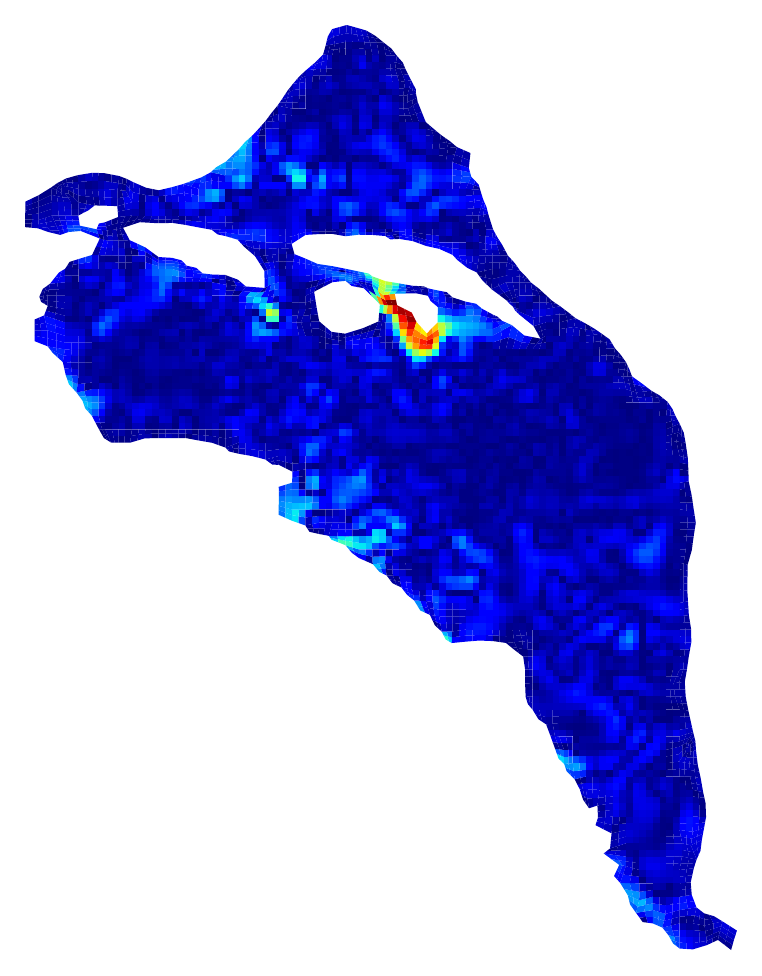} &
        \includegraphics[scale=0.25]{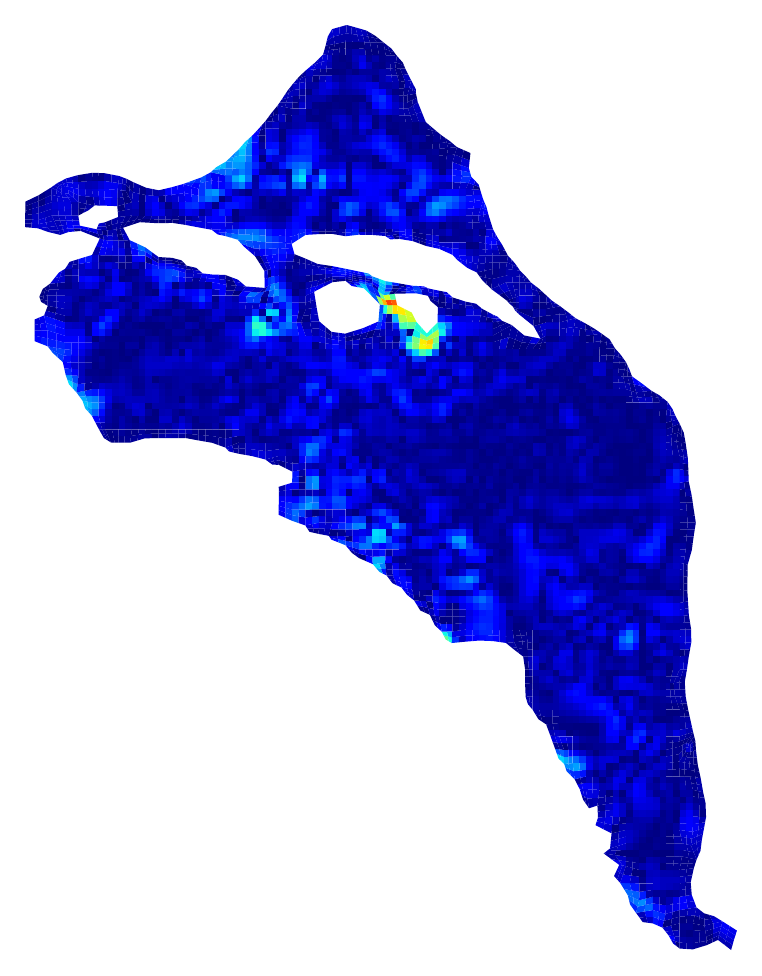} &
        \colorbar{0}{6}{{0,1,2,...,6}} \\
    \end{tabu}
    \caption{The fine-resolution ($N_{FV}=5900$) reference $y$ fields, the CKLEMAP and MAP estimates of the $y$ field and their point errors as functions of $N_{\mathbf{y}_{\mathrm{s}}}$.}\label{fig:hanford_1x}
\end{figure}

We start with the low-resolution model.
Figure~\ref{fig:hanford_1x} shows the locations of $y$ measurements, the $y$ fields estimated by the MAP and CKLEMAP methods for $N_{\mathbf{y}_\mathrm{s}} = 25, 50, 100$, and $200$, and the distributions of point errors in the MAP and CKLEMAP estimates of $y$ relative to the reference field $\tilde{y}$. 
For the considered measurement locations, we observe that the MAP and CKLEMAP methods have comparable accuracy for all $N_{\mathbf{y}_\mathrm{s}}$.

\cref{tab:1x_results} shows the ranges of relative $\ell_2 $ and absolute $\ell_\infty$ errors in the MAP and CKLEMAP $y$ estimates as well as the number of iterations in the minimization algorithm and the execution times (in seconds) for $N_{\mathbf{y}_{\mathrm{s}}}$ ranging from 25 to 200.
Also included in this table are the execution times of the accelerated CKLEMAP method. We note that the accuracy (including the $\ell_2 $ and absolute $\ell_\infty$ errors) and the number of iterations in the accelerated CKLEMAP and CKLEMAP methods are the same.

As expected, the accuracy of the MAP and CKLEMAP methods increases with $N_{\mathbf{y}_{\mathrm{s}}}$.
The MAP and CKLEMAP methods are almost equally accurate, with $\ell_2$ and $\ell_\infty$ errors in the CKLEMAP method being slightly smaller.
However, we observe that CKLEMAP is faster than MAP for all considered values of $N_{\mathbf{y}_\mathrm{s}}$ except for $N_{\mathbf{y}_\mathrm{s}} =25$, where the MAP's lower bound of the execution time is less than that of the CKLEMAP\@.
Accelerated CKLEMAP is about 20\% faster than CKLEMAP and for all considered values of $N_{\mathbf{y}_{\mathrm{s}}}$.
Accelerated CKLEMAP is also faster than MAP\@ for all considered cases; however, the speedup depends on $N_{\mathbf{y}_{\mathrm{s}}}$. 

In all examples reported in \cref{tab:1x_results}, the number of unknowns in the  CKLEMAP method is 1000 (the number of terms in the CKLE expansion), while in the MAP method, this number is 1475 (the number of cells in the FV model).
The reason for CKLEMAP being slower than MAP for $N_{\mathbf{y}_{\mathrm{s}}}=25$ and certain $y$ measurement locations is that for such locations MAP converges much faster.
For example, the lower execution time bands in MAP and CKLEMAP  correspond to 29 and 50 iterations, respectively.
However, because there are fewer unknowns in the CKLEMAP method, the CKLEMAP computational time per iteration is smaller than that in MAP\@.
As a result, the computational time in the CKLEMAP is only 20\% larger than that of MAP for these limiting cases.
The time per iteration is further reduced in the accelerated CKLEMAP method, resulting in the execution time of accelerated CKLEMAP being less than that of MAP by 20\%. 
We also note that for $N_{\mathbf{y}_{\mathrm{s}}}>25$, MAP requires more iterations than CKLEMAP, making the computational advantages of CKLEMAP even more significant.  

Next, we perform a similar study for the medium-resolution model with $N = 5900$ cells.
Table~\ref{tab:4x_results} provides a comparative summary of the models considered for this case.
Here, we find that CKLEMAP is slightly more accurate than MAP  for all considered values of $N_{\mathbf{y}_{\mathrm{s}}}$ and one to two orders of magnitude faster than MAP\@.
Accelerated CKLEMAP is approximately 10\% faster than CKLEMAP\@.
The computational advantage of CKLEMAP significantly increases with the problem size as the number of unknown parameters in the MAP linearly increases with the problem size while the number of parameters in the CKLEMAP is independent of the problem size.

\subsection{Scaling of the execution time with the problem size} 
\begin{table}[!htbp]
    \centering
    \caption{Performance of MAP and CKLEMAP in estimating the coarse-resolution (${N_{FV}=1475}$) mesh as functions of $N_{\mathbf{y}_{\mathrm{s}}}$.}\label{tab:1x_results}%
    \begin{tabu} to \textwidth {rX[1.3,rm]*{4}{X[cm]}}
        \toprule
        & & \multicolumn{4}{c}{$N_{\mathbf{y}_{\mathrm{s}}}$} \\
        & solver & 25 & 50 & 100 & 200 \\
        \midrule
        \multirow{2}{*}{\makecell[r]{least square\\iterations}} & MAP & 29--95 & 29--106 & 41--60 & 28--80 \\
        & CKLEMAP & 50--96 & 26--70 & 20--65 & 33--62 \\
        \midrule
        \multirow{3}{*}{\makecell[r]{execution\\time (s)}} & MAP & 31.36--91.50 & 57.98--199.61 & 76.39--123.08 & 32.88--80.32 \\
        & CKLEMAP & 37.01--71.04 & 21.71--51.86 & 14.25--36.29 & 17.73--40.57 \\
        & accelerated CKELMAP & 25.45--47.97 & 17.00--40.00 & 12.07--30.41 & 12.78--29.09 \\
        \midrule
        \multirow{2}{*}{\makecell[r]{relative\\$\ell_2$ error}} & MAP & 0.092--0.111 & 0.084--0.101 & 0.073--0.084 & 0.068--0.073 \\
        & CKLEMAP & 0.091--0.109 & 0.082--0.101 & 0.072--0.083 & 0.064--0.071 \\
        \midrule
        \multirow{2}{*}{\makecell[r]{absolute\\$\ell_\infty$ error}} & MAP & 5.38--6.61 & 4.95--6.55 & 4.06--6.35 & 3.88--6.74 \\
        & CKLEMAP & 4.96--6.25 & 4.73--6.11 & 3.46--5.31 & 5.63--5.71 \\
        \bottomrule
    \end{tabu}
\end{table}

\begin{table}[!htbp]
    \centering
    \caption{Performance of MAP and CKLEMAP in estimating the fine-resolution (${N_{FV}=5900}$) mesh as functions of $N_{\mathbf{y}_{\mathrm{s}}}$.}\label{tab:4x_results}%
    \begin{tabu} to \textwidth {rX[1.3,rm]*{4}{X[cm]}}
        \toprule
        & & \multicolumn{4}{c}{$N_{\mathbf{y}_{\mathrm{s}}}$} \\
        & solver & 25 & 50 & 100 & 200 \\
        \midrule
        \multirow{2}{*}{\makecell[r]{least square\\iterations}} & MAP & 78--99 & 71--97 & 69--83 & 23--76 \\
        & CKLEMAP & 53--114 & 20--142 & 36--60 & 15--83 \\
        \midrule
        \multirow{3}{*}{\makecell[r]{execution\\time (s)}} & MAP & 3907.00--4868.21 & 3528.90--4580.40 & 3533.06--4190.20 & 1247.37--3733.05 \\
        & CKLEMAP & 88.76--181.67 & 48.45--200.08 & 62.59--100.04 & 42.86--148.03 \\
        & accelerated CKELMAP & 77.05--141.90 & 39.50--156.63 & 52.14--81.19 & 38.28--120.18 \\
        \midrule
        \multirow{2}{*}{\makecell[r]{relative\\$\ell_2$ error}} & MAP & 0.0954--0.112 & 0.081--0.105 & 0.074--0.088 & 0.065--0.073 \\
        & CKLEMAP & 0.0906--0.111 & 0.081--0.105 & 0.068--0.079 & 0.061--0.069 \\
        \midrule
        \multirow{2}{*}{\makecell[r]{absolute\\$\ell_\infty$ error}} & MAP & 4.96--7.21 & 5.45--7.28 & 4.00--6.48 & 4.37--5.20 \\
        & CKLEMAP & 4.21--6.66 & 4.94--6.74 & 3.79--5.71 & 3.82--5.28 \\
        \bottomrule
    \end{tabu}
\end{table}

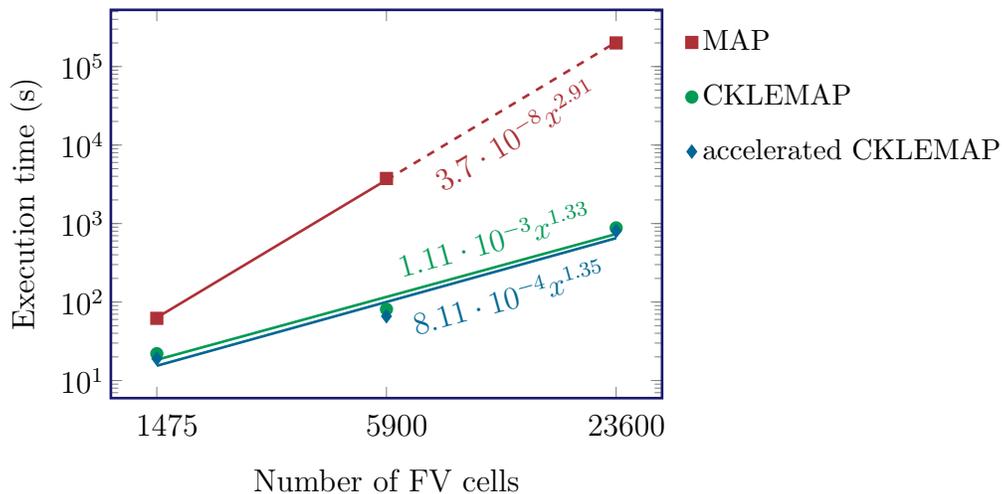
\begin{figure}[!htbp]
    \centering
    \begin{tikzpicture}
        \begin{loglogaxis}[trend plot]
            \pgfplotstableread{figures/time_vs_size.txt}{\tabledata}
            \newcounter{yi}
            \pgfplotsinvokeforeach {MAP_t, CKLEMAP_1000_t, CKLEMAP_acc} {
                \pgfmathsetmacro{\nodepos}{ifthenelse(\the\value{yi}==1,"above","below")}
                \setv{nodepos}{#1}{\nodepos}
                \pgfmathsetmacro{\pos}{".75"}
                \setv{pos}{#1}{\pos}
                \addplot+ table [x=cells, y=#1] {\tabledata};
                \label{plot:#1}
                \addplot+[mark=none] table [x=cells, y={create col/linear regression={y=#1}}] {\tabledata}
                    node[\getv{nodepos}{#1}, pos=\getv{pos}{#1}, sloped] {
                        $\pgfmathprintnumber{\getv{intercept}{#1}} x^{\pgfmathprintnumber{\getv{slope}{#1}}}$
                    }
                    coordinate[pos=0] (A_#1) coordinate[pos=.5] (B_#1) coordinate[pos=1] (C_#1);
                \setv{slope}{#1}{\pgfplotstableregressiona}
                \pgfkeys{/pgf/fpu=true}
                \pgfmathparse{exp(\pgfplotstableregressionb)}
                \setv{intercept}{#1}{\pgfmathresult}
                \pgfkeys{/pgf/fpu=false}
                \pgfmathaddtocounter{yi}{1}
            }
            \draw[Maroon] (A_MAP_t) -- (B_MAP_t);
            \draw[Maroon, dashed] (B_MAP_t) -- (C_MAP_t);
            \legend{MAP, , CKLEMAP, , accelerated CKLEMAP}
        \end{loglogaxis}
    \end{tikzpicture}
    \caption{Execution times of MAP, CKLEMAP, and accelerated CKLEMAP methods versus the number of FV cells. The execution times of MAP for the mesh with 23600 FV cells are estimated by extrapolation.}\label{fig:time}
\end{figure}

The comparison of Tables~\ref{tab:1x_results} and~\ref{tab:4x_results} shows that the execution times of the MAP, CKLEMAP, and accelerated CKLEMAP increase with the mesh resolution; however, the execution times of CKLEMAP and accelerated CKLEMAP increase slower than that of MAP\@.
To study the scalability of these methods with the problem size, we use these methods to estimate $y$ in the high-resolution FV model with $N = 23600$ and, in Figure~\ref{fig:time}, we plot the execution times of these methods as functions of $N$.
The number of $y$ measurements in all simulations reported in this figure is set to $N_{\mathbf{y}_{\mathrm{s}}}=100$.
We also show the power-law models fitted to the scalability curves computed using MAP, CKLEMAP, and accelerated CKLEMAP\@.
We note that for $N = 23600$, the MAP method did not converge after running for two days.
Therefore, the power law relationship for the MAP method is obtained based on the execution times for $N = 1475$ and $5900$ and used to estimate the MAP's execution time for the highest resolution by extrapolation. 
We find that the MAP, CKLEMAP, and accelerated CKLEMAP execution times scale as $N^{2.91}$, $N^{1.33}$, and $N^{1.35}$, respectively.
Therefore, the CKLEMAP methods have a computational advantage over the MAP method for large problems.
The CKLEMAP and accelerated CKLEMAP methods have approximately the same scalability, but for the same problem size, the accelerated CKLEMAP method is 10--20\% faster than the CKLEMAP method.   

\section{Discussion and Conclusions}\label{sec:conclusions}

We proposed the CKLEMAP method as an alternative to the MAP methods for solving inverse PDE problems and used it for estimating the transmissivity and hydraulic head in a two-dimensional steady-state groundwater model of the Hanford Site.
The CKLEMAP method is based on the approximation of unknown parameters (log-transmissivity in this case) with CKLEs.
The advantage of using a CKLE over other representations (like DNNs in~\cite{xu-2021-CMAME}) is that it enforces (i.e., exactly matches) the field measurements and the covariance structure, that is, it models the field as a realization of the conditional Gaussian field with a prescribed covariance function. 
As a general conclusion, we found that the accuracy of the MAP and CKLEMAP methods is essentially the same (with CKLEMAP being a few percents more accurate under most tested conditions), but CKLEMAP is faster than MAP\@.

Specifically, we demonstrated that the CKLEMAP and MAP execution times scale with the problem size as $N^{1.33}$ and $N^{2.91}$, respectively, where $N$ is the number of FV cells.
The close-to-linear scaling of CKLEMAP's execution time with problem size gives CKLEMAP a computational advantage over the MAP method for large-scale problems.
We consider this to be the main advantage of the CKLEMAP method. 

For the same number of measurements, the accuracy of MAP and CKLEMAP can depend on the measurement locations.
Both the MAP and the CKLEMAP methods are, on average, equally accurate in terms of absolute $\ell_\infty$ errors. The CKLEMAP method is slightly more accurate than the MAP method in terms of relative $\ell_2$ errors.
The execution times of MAP and CKLEMAP increase, and their accuracy decreases, as the number of $y$ measurements decreases.

In the CKLEMAP method, execution time and accuracy increase with the increasing number of CKL terms.
In this work, as a baseline, we used $N_y = 1000$, which corresponds to $\text{rtol} < 10^{-8}$. We stipulate that this criterion is sufficient to obtain a convergent estimate of $y$ with respect to the number of CKL terms.
 
To further reduce the computational time, we proposed the accelerated CKLEMAP method, which takes advantage of the sparse structure of the stiffness matrix in the FV discretization of the residual term.
We demonstrated that the scalability of the accelerated CKLEMAP and CKLEMAP methods is approximately the same; however, for the same problem size, accelerated CKLEMAP is 10--20\% faster than the CKLEMAP method.  

\section{Acknowledgments}
This research was partially supported by the U.S. Department of Energy (DOE) Advanced Scientific Computing program and the United States Geological Survey. Pacific Northwest National Laboratory is operated by Battelle for the DOE under Contract DE-AC05-76RL01830.  The data and codes used in this paper are available at \url{https://github.com/yeungyh/cklemap.git}.

\bibliographystyle{elsarticle-num}
\bibliography{cklemap}

\end{document}